\documentclass[runningheads]{llncs}

\usepackage{eccv}

\usepackage{eccvabbrv}

\usepackage{graphicx}
\usepackage{booktabs}
\usepackage{amssymb}
\usepackage{enumitem}

\usepackage[accsupp]{axessibility}  

\usepackage[pagebackref=true,colorlinks,citecolor=brown]{hyperref}
\usepackage{adjustbox}
\usepackage{bm}
\usepackage{multirow}
\usepackage{multicol}
\usepackage{color, colortbl}
\usepackage{pifont}
\usepackage{placeins}
\usepackage{transparent}
\usepackage{algorithm}
\usepackage{algorithmic}
\usepackage{balance}
\usepackage{xcolor}

\usepackage[capitalize]{cleveref}
\Crefname{section}{Section}{Sections}
\crefname{section}{Sec.}{Secs.}
\Crefname{align}{Equation}{Equations}
\crefname{align}{Eq.}{Eqs.}
\Crefname{equation}{Equation}{Equations}
\crefname{equation}{Eq.}{Eqs.}
\Crefname{figure}{Figure}{Figures}
\crefname{figure}{Fig.}{Figs.}
\Crefname{table}{Table}{Tables}
\crefname{table}{Tab.}{Tabs.}

\newcommand\minisection[1]{\vspace{1mm}\noindent \textbf{#1}}

\definecolor{Gray}{gray}{0.9}
\definecolor{Celadon}{rgb}{0.67, 0.88, 0.69}
\definecolor{Cream}{rgb}{1.0, 0.99, 0.82}

\newcommand{\cmark}{{\color{green!70!black}\ding{51}}}
\newcommand{\xmark}{{\color{red}\ding{55}}}

\makeatletter
\newcommand*{\@rowstyle}{}
\newcommand*{\rowstyle}[1]{
  \gdef\@rowstyle{#1}%
  \@rowstyle\ignorespaces%
}
\newcolumntype{=}{
  >{\gdef\@rowstyle{}}%
}
\newcolumntype{+}{
  >{\@rowstyle}%
}
\makeatother

\usepackage{orcidlink}

\begin{document}
\newcommand{\NAME}{\textsc{MemoBench}\xspace}
\title{\NAME: Benchmarking World Modeling in Dynamically Changing Environments} 

\titlerunning{MemoBench}


\author{\normalfont
\begin{minipage}{\textwidth}\centering
Haoyu Chen\textsuperscript{1} \enspace
Kaichen Zhou\textsuperscript{1,2} \enspace
Hang Hua\textsuperscript{3} \enspace
Kaile Zhang\textsuperscript{4} \enspace
Jingwen Qian\textsuperscript{5} \\
Wufei Ma\textsuperscript{6} \enspace
Haonan Chen\textsuperscript{1} \enspace
Chunjiang Liu\textsuperscript{7} \enspace
Yizhou Zhao\textsuperscript{7} \enspace
Xiaoyuan Wang\textsuperscript{7} \\
Weiyue Li\textsuperscript{1} \enspace
Alan Yuille\textsuperscript{6} \enspace
Paul Pu Liang\textsuperscript{2} \enspace
Yilun Du\textsuperscript{1,8}
\end{minipage}
}

\authorrunning{H. Chen et al.}

\institute{\normalfont
\begin{minipage}{\textwidth}\centering
\textsuperscript{1}Harvard University \quad
\textsuperscript{2}MIT \quad
\textsuperscript{3}MIT-IBM Watson AI Lab \quad
\textsuperscript{4}Boston University \quad
\textsuperscript{5}Google\quad
\textsuperscript{6}JHU\quad
\textsuperscript{7}CMU\quad
\textsuperscript{8}Kempner Institute \quad
\end{minipage}
}

\maketitle
\begin{abstract}
    Video generation models aspire to simulate dynamic environments,
    and several benchmarks now evaluate memory consistency across
    frames. However, most assess consistency only while the target
    remains in view, and the few that force objects out of view
    evaluate static scenes where nothing changes during occlusion. 
    To bridge this gap, we introduce \NAME, a diagnostic benchmark built around the
    disappear-and-reappear paradigm in dynamically changing
    environments: a target object undergoes a physical process,
    disappears from view, and must be
    correctly recovered in its updated state upon reappearance. We curate 360 ground-truth clips spanning synthetic and real-world scenes, and design an evaluation suite combining automated metrics with
    VQA-based assessment across four diagnostic pillars.
    Evaluation of ten state-of-the-art models reveals key insights and open
    challenges regarding memory consistency under the
    disappear-and-reappear paradigm. Our dataset, code, and leaderboard are available at \url{https://github.com/MemoBench-Team}.
    
\end{abstract}
\begin{figure*}[h]
\centering
\includegraphics[width=0.95\textwidth]{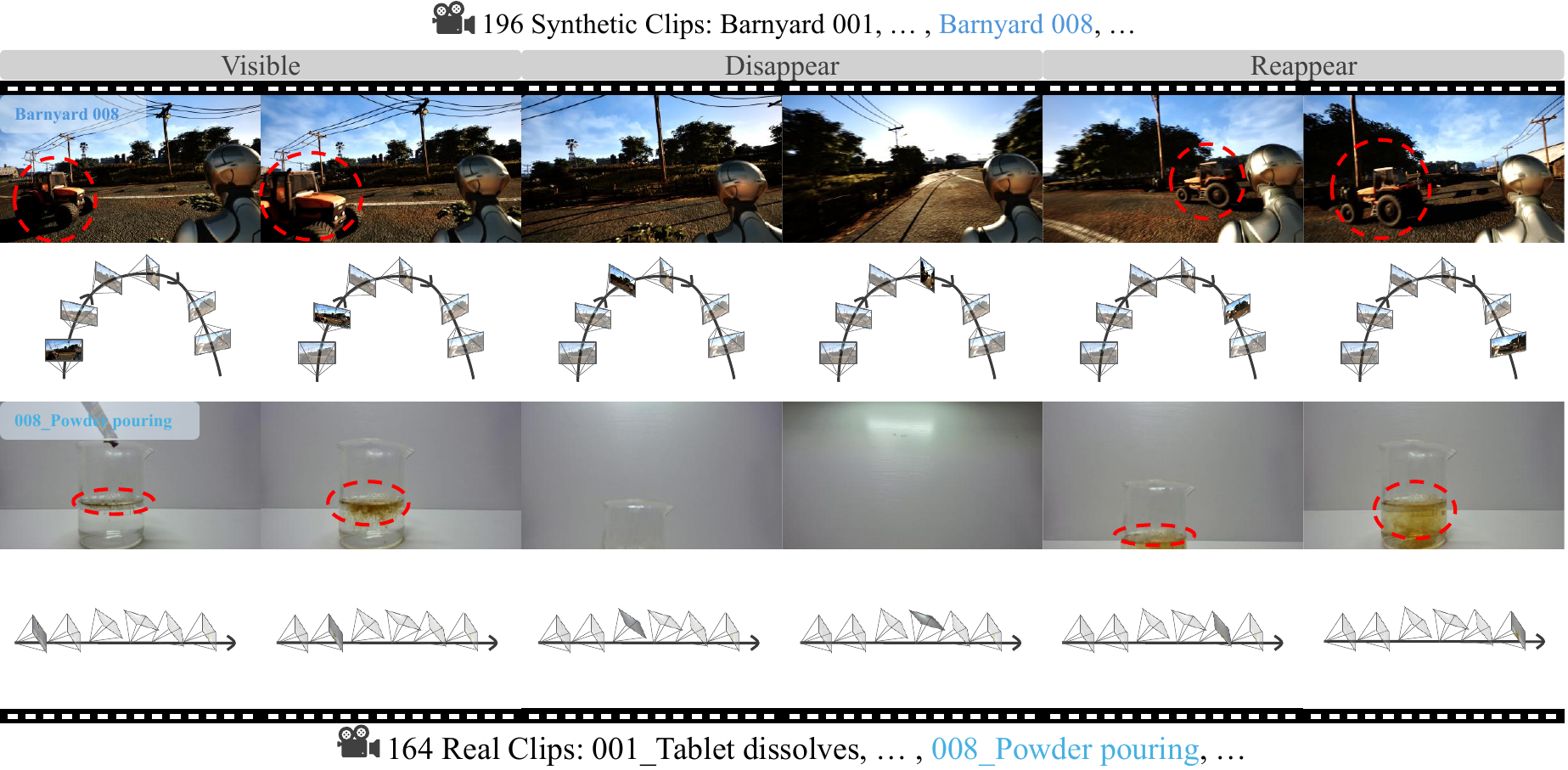}
\caption{
  \textbf{Overview of \NAME.}
  Rows~1--2 show a synthetic Visible--Disappear--Reappear sequence and
  its camera trajectory; Rows~3--4 show a real-world state-change
  sequence (powder pouring).
  \NAME contains 196 synthetic and 164 real-world clips, evaluated
  with automated metrics and LLM-judged VQA.
}
\label{fig:teaser}
\end{figure*}
\section{Introduction}
\label{sec:intro}
The real world is inherently dynamic, continuously evolving
regardless of whether anyone is watching: ice melts, flames
flicker, pedestrians walk, and traffic flows. Faithfully modeling such dynamically changing environments is
crucial to applications ranging from autonomous driving and robotic
manipulation to embodied tasks, where an agent must reason about how
the world has changed beyond its field of view. Recent progress
in video generation~\cite{zheng2024open,peng2025open} has shown that generative models can serve as \emph{world generators},
capturing environment dynamics and enabling prediction under actions or
interventions~\cite{ha2018world,hafner2023mastering,lingbot-world}.

Despite this ambition, a fundamental challenge remains under-explored:
\emph{visual memory under partial observability}. In cognitive science,
\textbf{object permanence}, the understanding that objects continue to
exist when out of sight, is among the earliest cognitive milestones.
An analogous capability is crucial for video generation: as the virtual
camera moves, objects inevitably leave and re-enter the field of view,
and the generative model must faithfully reproduce their appearance, position, and
any ongoing state changes upon return~\cite{lillemark2026flow}. This
\textbf{disappear-and-reappear} pattern is ubiquitous in everyday
experience. Yet current video generation
benchmarks seldom treat this as an explicit evaluation target, leaving
it unclear whether generative models truly \emph{remember} or merely
\emph{regenerate} scene content.

Existing benchmarks have advanced the evaluation of world generation along
multiple axes, including visual quality, temporal coherence, physical adherence,
and scene consistency~\cite{huang2024vbench,bansal2024videophy,li2025worldmodelbench,duan2025worldscore}, but
they predominantly evaluate what is \emph{continuously visible} across
frames. To our knowledge, none directly tests whether a generative model can maintain and update the
state of objects that have temporarily left the field of view, under
simultaneous camera and scene dynamics, leaving
it unclear whether models can preserve identity, geometry, and evolving
physical state across periods of occlusion.

To fill this gap, we introduce \NAME, a simple yet comprehensive diagnostic
benchmark for \emph{world modeling in dynamically changing environments}.
Each example follows a disappear-and-reappear structure:
(i)~the target object is \emph{visible} and undergoing a physical process;
(ii)~the camera pans away and the target \emph{disappears} from view while
the process continues naturally;
and (iii)~the camera returns and the target \emph{reappears}, and the generative model
must recover its updated state.As illustrated in Fig.~\ref{fig:teaser}, \NAME includes both synthetic and
real-world examples following this Visible--Disappear--Reappear paradigm, together with camera trajectories and a comprehensive evaluation setup.
\NAME provides camera
trajectories and depth maps, enabling evaluation grounded in both geometry and physical state evolution.

Our contributions are as follows:
\begin{itemize}
    \item We introduce \NAME, the first benchmark that evaluates
        memory consistency in world generation through the
        disappear-and-reappear paradigm, comprising 360
        high-quality ground-truth videos at 1920$\times$1080 resolution spanning
        diverse scenes and physical-state changes.
    \item We design a comprehensive evaluation suite combining
        automated metrics (video quality, Object Reappearance
        Score, pixel-level fidelity, and camera controllability)
        with LLM-judged VQA across four diagnostic dimensions.
    \item We benchmark ten state-of-the-art world generation
        models, revealing that no current model reliably
        maintains object memory across occlusion, and identifying
        key open challenges for future work.
\end{itemize}
\section{Related Work}

\minisection{Video Generation as World Simulation.}
Video generation has evolved from synthesizing short clips to world simulation, modeling physics, causal dynamics, and persistent state of environments \cite{ha2018world,hafner2023mastering,ho2022video}.
A large body of work has pushed the boundary of realistic video synthesis~\cite{chen2024videocrafter2,li2026comprehensive, chen2023motion,zhao2025total,chen2023seine,he2022latent,lin2024open,luo2023videofusion,singer2022make,wang2023modelscope,wang2025lavie,liu2024emo,xiang2024pandora,zhao2025masiv,liu2026mosiv,xing2024dynamicrafter,zhang2025show,zheng2024open,openai_sora,peng2025open, openai_sora2,kuaishou_kling,luma2024, runway2024}, with notable models including
CogVideoX~\cite{yang2024cogvideox}, Open-SoRA~\cite{zheng2024open}, LTX-Video~\cite{hacohen2024ltx} and LingBot-World~\cite{lingbot-world} which explicitly targets long-term memory.
Despite these advances, it remains unclear how well today’s models preserve a persistent world state, rather than generating visually convincing frames.

\minisection{Camera-Controllable Video Generation.}
A critical step toward faithful world simulation is generating videos conditioned on explicit camera trajectories, enabling controlled traversal of 3D environments. Early methods~\cite{he2024cameractrl,wang2024motionctrl} introduced camera pose conditioning modules for diffusion models, with later methods~\cite{xu2024camco, zheng2024cami2v, yu2024viewcrafter,voleti2024sv3d,wang2025holigs,zhao2026geostream,liu2026omniroam,lin2026depth,ge2026airsim360,liu2026driveva,liu20264dstr,zhou2026gem,zhou2026stream3d} improving camera control, 3D consistency, and geometry-aware panoramic data construction through multi-view constraints, explicit 3D representations, geometric conditioning, and simulation.
Recent control-conditioned text-image-to-video (C+TI2V) models~\cite{lingbot-world,wan2025wan,dai2025fantasyworld,hyworld2025,li2025hunyuan}, where the control signal is either an explicit camera pose sequence or an action input, allow precise viewpoint control essential for autonomous driving and embodied AI.
This architectural distinction carries important evaluation implications: C+TI2V models can execute trajectories that move objects in and out of view, whereas TI2V models may exhibit \emph{inactivity}, trivially satisfying visual consistency checks by keeping the viewpoint mostly static.

\begin{table*}[t]
\centering
\caption{
\textbf{Comparison with recent world generation benchmarks.}
Scene Trav.\ = spatial traversal within a generated sequence;
Phys.\ Adh.\ = physical adherence;
Obj.\ Perm.\ = object permanence via disappear-and-reappear.
\NAME is the only benchmark that explicitly evaluates memory
consistency through the disappear-and-reappear paradigm.
}
\setlength{\tabcolsep}{0.5em}
\renewcommand{\arraystretch}{1.15}
\begin{adjustbox}{width=\linewidth}
\begin{tabular}{l c c cccccc}
\toprule
\textbf{Benchmark}
  & \shortstack{\#\,Eval.}
  & \shortstack{Eval.\\Type}
  & \shortstack{Scene\\Trav.}
  & \shortstack{Long\\Seq.}
  & \shortstack{Camera\\Ctrl.}
  & \shortstack{Scene\\Consist.}
  & \shortstack{Phys.\\Adh.}
  & \shortstack{Obj.\\Perm.} \\
\midrule
VideoPhy-2~\cite{bansal2025videophy}
  & 3{,}940 & Text
  & \xmark & \xmark & \xmark & \xmark & \cmark & \xmark \\
WorldModelBench~\cite{li2025worldmodelbench}
  & 350 & Text+Img
  & \xmark & \xmark & \xmark & \xmark & \cmark & \xmark \\
WorldSimBench~\cite{qin2024worldsimbench}
  & 2{,}831 & Text+Img
  & \xmark & \xmark & \xmark & \xmark & \cmark & \xmark \\
World-in-World~\cite{zhang2025world}
  & 1{,}079 & Episode
  & \cmark & \cmark & \cmark & \xmark & \xmark & \xmark \\
VBench~2.0~\cite{huang2024vbench}
  & 946 & Text
  & \cmark & \xmark & \cmark & \cmark & \cmark & \xmark \\
WorldScore~\cite{duan2025worldscore}
  & 3{,}000 & Img+Traj
  & \cmark & \cmark & \cmark & \cmark & \xmark & \xmark \\
\midrule
\rowcolor[HTML]{ECF4FF}
\textbf{MemoBench (Ours)}
  & 360 & Img+Traj+GT Video
  & \cmark & \cmark & \cmark & \cmark & \cmark & \cmark \\
\bottomrule
\end{tabular}
\end{adjustbox}
\label{tab:bench_compare}
\end{table*}

\minisection{Evaluation Benchmarks for Video Generation.}
Most evaluation benchmarks evaluate video quality through dimensional decomposition
~\cite{huang2024vbench,liu2023fetv,yuan2024chronomagic,sun2025t2v} or physical adherence ~\cite{bansal2024videophy,bansal2025videophy,kang2024far}, while recent work evaluates generators as world simulators ~\cite{li2025worldmodelbench,qin2024worldsimbench}. However, these all evaluate single-viewpoint clips and to our knowledge none tests whether models maintain world state when previously observed content reappears.
Recent work has made progress. WorldScore~\cite{duan2025worldscore} evaluates scene consistency across multi-view sequences constrained by camera trajectories, but does not test object permanence. World-in-World~\cite{zhang2025world} evaluates world models in closed-loop embodied settings, focusing on task-level success rather than fine-grained visual consistency of individual objects.

As summarized in \cref{tab:bench_compare}, these benchmarks collectively
advance the evaluation of scene traversal, camera control, and scene
consistency, yet they do not address the joint challenge of \emph{dynamic
camera viewpoints} and \emph{dynamic scene content}, which tests whether a model
can maintain the evolving state of a target object after it disappears
from view and correctly recover it upon reappearance.
Our \NAME fills this gap through the disappear-and-reappear
paradigm, requiring models to maintain memory of objects that leave the
field of view and recover their evolved state when they reappear, directly
probing memory consistency under simultaneous camera and scene motion.
\section{\NAME}
\subsection{Data Curation Framework}
\label{sec:data-curation}

\begin{figure*}[t]
  \centering
  \includegraphics[width=\textwidth]{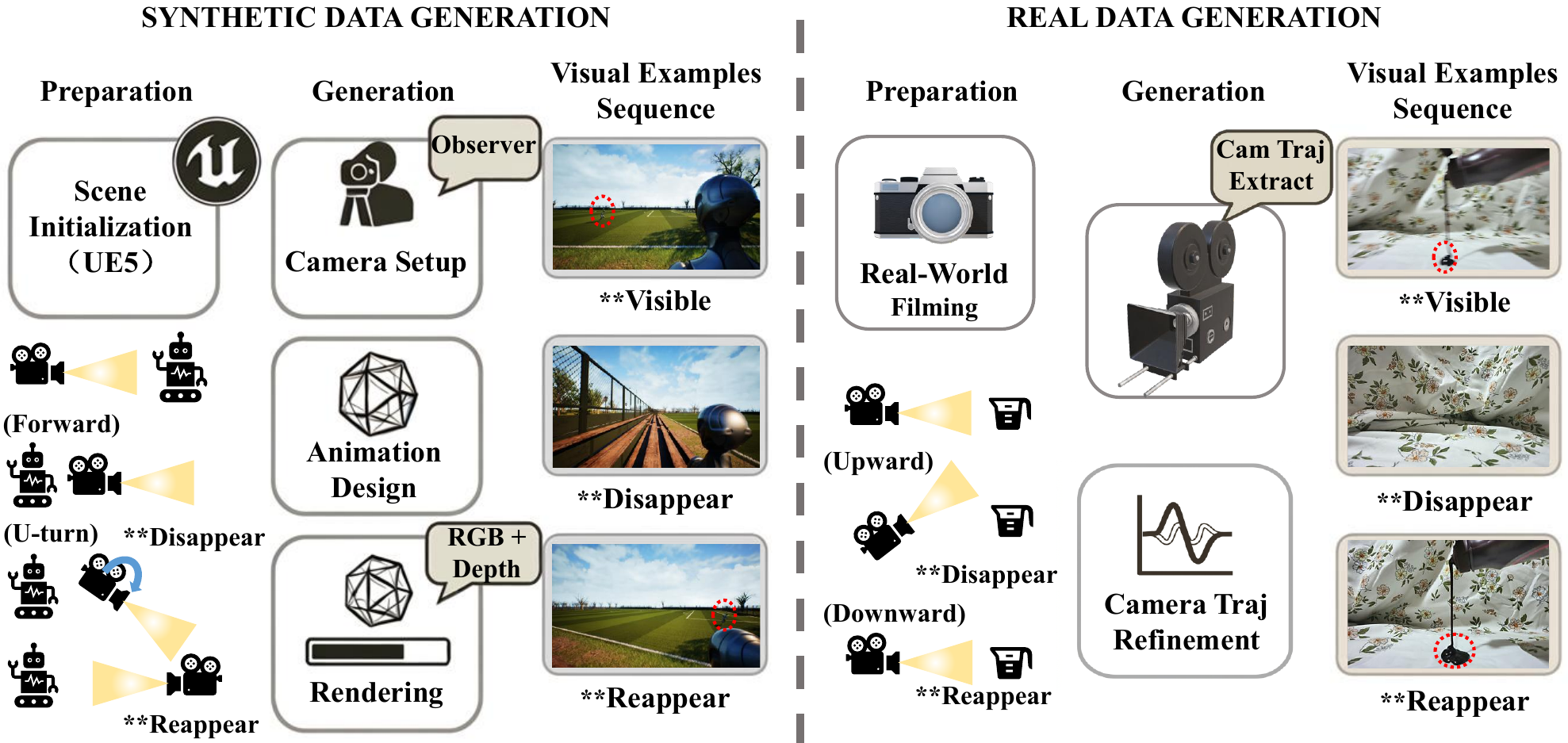}
  \caption{\textbf{Data curation pipeline for \NAME.} Left: synthetic data (196 clips, 14 scene subdomains across 5 environment categories) generated in Unreal Engine~5. Right: real-world data (164 clips, 30 physical-state-change processes across 7 categories) captured in controlled indoor settings.}
  \label{fig:pipeline}
\end{figure*}

Our data curation pipeline comprises two parallel workflows
as illustrated in~\cref{fig:pipeline}: a synthetic pipeline and a
real-world pipeline.

\noindent\underline{Synthetic pipeline.}
We initialize diverse 3D scenes in Unreal Engine~5 and place animated
target objects along predefined paths. A virtual camera is attached to a first-person observer who follows a scripted trajectory: the observer first faces the target (Visible),
performs a head turn or U-turn that moves the target out of the field
of view (Disappear), and continues along the trajectory until the
target re-enters the frame (Reappear).
Each clip is rendered at 1920$\times$1080 (60\,FPS) with per-frame
RGB, metric depth, camera intrinsics, and camera-to-world poses
exported automatically.

\noindent\underline{Real-world pipeline.}
We record diverse physical-state-change processes in controlled indoor settings using a fixed-position camera that pans away from the target object and then returns, creating the same three-phase structure. Camera intrinsics are obtained from manufacturer calibration, while extrinsic poses are estimated from the recorded RGB frames using MapAnything~\cite{keetha2025mapanything}, followed by trajectory smoothing to obtain clean per-frame camera-to-world poses.

\subsection{Dataset Overview}
\label{sec:dataset-overview}
\NAME comprises 360 ground-truth video clips organized into two
complementary subsets. The synthetic subset (196 clips) focuses on
\emph{spatial diversity}, spanning 14 scene subdomains across five
environment categories with rich ego-motion driving the
disappear-and-reappear structure. The real-world subset (164 clips)
focuses on \emph{material diversity}, covering 30 physical-state-change
processes across seven categories that depend on properties such as
viscosity, elasticity, and thermal conductivity, which game engines
cannot accurately model. Dataset statistics and breakdowns are provided 
in the supplementary material \cref{fig:dataset_overview}.

Each clip is human-annotated with two keyframe indices:
$d_{\mathrm{start}}$ (the frame at which the target has completely
disappeared from the FOV) and $r_{\mathrm{start}}$ (the frame at
which the target has fully reappeared), which are linearly
mapped to the generated-video length to set the
disappear-and-reappear interval for evaluation.

\subsection{Evaluation Setup}
\label{sec:eval-setup}
Given an input reference image (the first frame), a text prompt,
and optionally a camera-control signal, a generative model produces
a short video of $T$ frames (see supplementary \cref{tab:model_config} for per-model configurations). Since the ground-truth (GT) and generated
videos may differ in frame count and frame rate, GT frames are uniformly
downsampled by linearly interpolating frame indices before computing
per-frame metrics.

We report two complementary evaluations:
~\textbf{(1) Automated metrics} (\cref{sec:auto-metrics}) computed directly from the generated and GT videos;
~\textbf{(2) VQA-based evaluation} (\cref{sec:vqa-metric}) using Yes/No questions grouped into diagnostic dimensions.

\subsection{Automated Metrics}
\label{sec:auto-metrics}

\minisection{Phase Structure.}
Using the two annotated keyframes defined in \cref{sec:eval-setup},
each evaluation clip is divided into three non-overlapping phases:
Visible~(V): frames $[0, d_{\mathrm{start}})$, where the target is fully in view;
Disappeared~(D): frames $[d_{\mathrm{start}}, r_{\mathrm{start}})$, where the target is completely out of view;
Reappear~(R): frames $[r_{\mathrm{start}}, N{-}1]$, where the target has fully re-entered the field of view.
Unless specified, we exclude the D phase from motion and geometry metrics by design.

\minisection{Normalization to a 0--100 Scale.}
Each raw metric $m$ is mapped to a percentage score via clipped min--max normalization:
\begin{equation}
\mathcal{N}(m; a,b)
= 100 \cdot
\mathrm{clip}_{[0,1]}
\!\left(\frac{m-a}{b-a}\right),
\label{eq:norm}
\end{equation}
where $[a,b]$ is a predefined valid range for that metric and $\mathrm{clip}_{[0,1]}(\cdot)$ denotes truncation to the interval $[0,1]$.
We use fixed ranges for all composite metrics (e.g., Aesthetic in $[1,10]$; CLIP-IQA+ in $[0,1]$).

\minisection{General Video Quality.}
We report four metrics: Visual Quality, Motion Smoothness, Object Identity Consistency, and Geo3D Consistency.

\noindent\underline{Visual Quality.}
We average two no-reference quality signals over uniformly sampled frames from all phases:
(i)~\emph{AestheticScore} from the LAION aesthetic predictor~\cite{schuhmann2022laion} (${\sim}0$--$10$);
(ii)~\emph{ImageQuality} from CLIP-IQA+~\cite{wang2023exploring} ($[0,1]$).
Both are mapped to $[0,100]$ via Eq.~\ref{eq:norm} and averaged:
\begin{equation}
S_{\mathrm{vq}} = \frac{1}{2}\bigl(\mathcal{N}(s_a;\,1,10) + \mathcal{N}(s_q;\,0,1)\bigr),
\end{equation}
where $s_a$ denote the mean AestheticScore over sampled frames and $s_q$ denote the mean CLIP-IQA+ score over sampled frames.

\noindent\underline{Motion Smoothness.}
We follow VBench~\cite{huang2024vbench} and use RAFT-Large~\cite{teed2020raft}
optical flow to measure temporal smoothness.
For consecutive sampled frame pairs within the V and R phases, RAFT predicts
a dense flow from frame $i$ to $i{+}1$, we warp frame $i$ with bilinear
sampling, and compute the mean L1 photometric error $\bar{e}$.
We define:
\begin{equation}
S_{\mathrm{ms}} = \mathcal{N}\!\left(\exp\!\left(-\frac{\bar{e}}{\tau}\right);\; 0,\; 1\right),
\end{equation}
where $\tau = 0.15$ is a temperature parameter. Lower warp error implies
smoother motion; the D phase is excluded by design.

\noindent\underline{Object Identity Consistency.}
We use DINOv2 ViT-B/14~\cite{oquab2023dinov2} patch tokens to measure
foreground object stability across the reappearance phase.
For each sampled R-phase frame, we compute per-patch cosine similarity
against the generated first frame $I_0$ patch tokens, and select the
top-$k$\% most similar patches ($k{=}40$) to focus on the persistent
foreground object.
Let $\bar{c}^{(t)}_{\mathrm{top}}$ and $c^{(t),\min}_{\mathrm{top}}$
denote the mean and minimum similarity over the top-$k$\% patches for
frame $t$. We aggregate across all sampled R-phase frames:
\begin{equation}
S_{\mathrm{oc}} = \alpha \cdot \overline{\bar{c}_{\mathrm{top}}}
+ (1 - \alpha) \cdot \min_t\, c^{(t),\min}_{\mathrm{top}},
\end{equation}
where $\overline{\bar{c}_{\mathrm{top}}}$ is the mean of per-frame
top-$k$\% means, $\min_t\, c^{(t),\min}_{\mathrm{top}}$ is the global
minimum across all sampled frames, and $\alpha = 0.7$.

\noindent\underline{Geo3D Consistency.}
Motion smoothness relies on optical flow, which captures pixel-level displacement but is sensitive to large camera motions and occlusions. To assess whether the underlying scene structure remains consistent, we compare per-frame depth maps estimated by Depth Anything V2~\cite{yang2024depth}. High cosine similarity between consecutive depth maps indicates stable 3D geometry, while low similarity reveals artifacts such as depth collapse or scene drift.
Each depth map is min--max normalized to $[0,1]$, flattened, and L2-normalized.
We compute cosine similarity between consecutive depth maps within the
V and R phases separately, obtaining per-phase mean ($\bar{d}$) and
minimum ($d^{\min}$) similarities:
\begin{equation}
S_{\mathrm{gc}} = \alpha \cdot
\frac{\bar{d}_V + \bar{d}_R}{2}
+ (1 - \alpha) \cdot
\min(\,d^{\min}_V,\; d^{\min}_R\,),
\end{equation}
where $\alpha = 0.7$. The D phase is excluded by design.

\minisection{Memory-Specific Metrics.}
We report five metrics across three groups: Object Reappearance Score (ORS), Pixel-Level Fidelity including PSNR, SSIM, and LPIPS, and Camera Controllability.

\noindent\underline{Object Reappearance Score (ORS).}
A key requirement of our evaluation is verifying whether the target object
reappears during the R phase. Because the camera viewpoint in the
R phase generally differs from the V phase (especially in synthetic clips
with free camera trajectories), spatial metrics such as mask IoU between
phases are unreliable. We therefore adopt a detection-based approach
using SAM-3~\cite{carion2025sam}, a text-prompted segmentation model.

For each R-phase frame, we query SAM-3 with the target object's text
description and apply coverage filtering (0.05\%--50\% of image area,
with a 0.05\%--70\% fallback) to reject spurious large-area masks
(e.g., robot body) and noise. A frame is considered a detection
if at least one valid mask is returned, and we record the highest
confidence score among valid masks. ORS is defined as:
\begin{equation}
S_{\mathrm{ors}} = \frac{n_d}{n_R} \cdot \frac{1}{n_d}\sum_{i=1}^{n_d} p_i,
\end{equation}
where $n_R$ is the total number of R-phase frames, $n_d$ is the number of frames with a valid detection, and $p_i$ is the confidence score of the $i$-th detected frame.
A high ORS indicates the model reliably regenerates a recognizable target
object when it reappears; a low ORS suggests the object is absent,
unrecognizable, or blended into the background.

\noindent\underline{Pixel-Level Fidelity.}
For clips where a ground-truth reference video is available, we compute
per-frame pixel-level fidelity between the generated and GT frames.
We report three complementary metrics per phase:
PSNR~\cite{hore2010image}~($\uparrow$) measuring signal fidelity;
SSIM~\cite{wang2004image}~($\uparrow$) measuring structural similarity;
and LPIPS~\cite{zhang2018unreasonable}~($\downarrow$) measuring perceptual
distance using a VGG backbone.
Scores are computed separately for the V, D, and R phases as well as the
full video (V+D+R), allowing phase-level analysis of where fidelity
degrades. In our main results we report whole-video averages.

\noindent\underline{Camera Controllability.}
We estimate per-frame camera-to-world poses from generated frames using 
MapAnything~\cite{keetha2025mapanything}, a feed-forward pose estimator
that scales to large evaluation without multi-view optimization~\cite{zhou2025page}, and align
the estimated trajectory to the GT via the first frame. We evaluate rotation error only, as the
disappear-and-reappear paradigm is driven by camera heading changes and
monocular translation is scale-ambiguous. We define:
\begin{equation}
S_{\mathrm{cc}} = \mathrm{clip}_{[0,1]}\!\left(1 - \frac{E_{\mathrm{rot}}}{\max(\Theta_{\mathrm{gt}},\;\theta_0)}\right),
\end{equation}
where $E_{\mathrm{rot}}$ is the ATE rotation RMSE (degrees) after
first-frame alignment, $\Theta_{\mathrm{gt}}$ is the end-to-end net
GT rotation, and $\theta_0 = 10^{\circ}$ prevents instability when
the camera returns close to its starting orientation.

\minisection{Prompt Fidelity.}
We report one metric: ImageReward Score.

\noindent\underline{ImageReward Score.}
We compute ImageReward~\cite{xu2023imagereward} on uniformly sampled
frames paired with the prompt. Raw scores (${\sim}{-}2$ to ${+}2$) are
first mapped to $[0,1]$ via sigmoid, then normalized to $[0,100]$ via
Eq.~\ref{eq:norm} with $[a,b]=[0,1]$.

\subsection{VQA-based Metrics}
\label{sec:vqa-metric}

\begin{figure}[t]
    \centering
    \includegraphics[width=\linewidth]{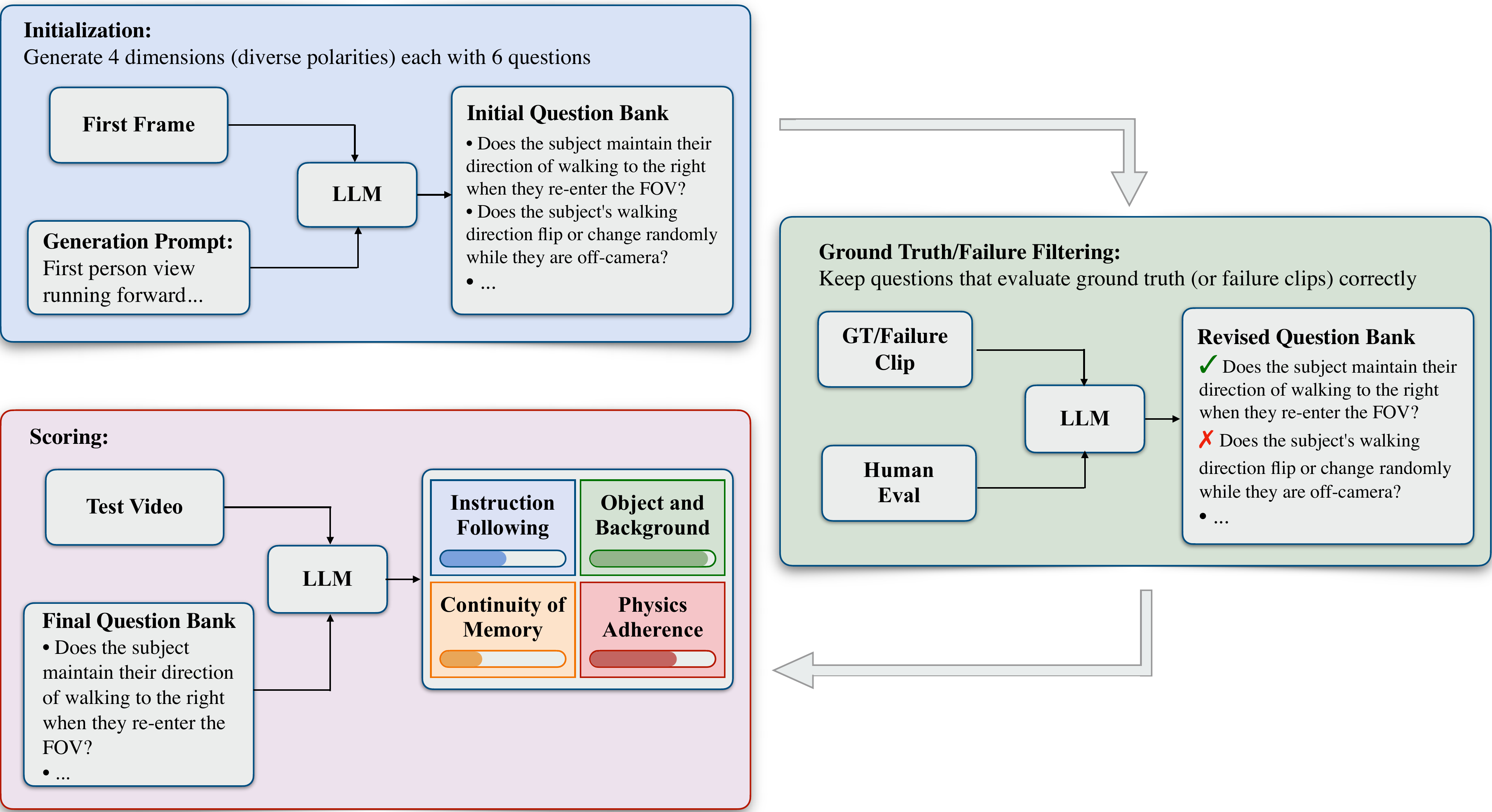}
    \caption{\textbf{VQA evaluation pipeline.} An LLM generates 24 polarity-balanced Yes/No questions (6 per dimension) from the prompt and first frame. Questions are filtered through ground-truth and failure-clip evaluation, then validated by human reviewers. The final question bank is applied to each generated video, producing per-dimension pass rates across four diagnostic dimensions.}
    \label{fig:vqa_pipeline}
\end{figure}

\minisection{Pipeline.}
Automated metrics primarily capture pixel-level fidelity and
low-level perceptual quality, but often fail to measure whether
a generated video correctly follows the prompt, maintains object
identity over time, or preserves physical plausibility.
Our VQA-based metric is designed to complement these automated
signals by evaluating higher-level semantic correctness and
temporal reasoning.

Recent work~\cite{lin2024evaluating,wang2026panoworld,hua2024mmcomposition,hua2024finematch,hu2023tifa,hu2023promptcap,ma20253dsrbench,feng2026visual,li2026grading,zheng2023judging}
shows that VQA-based evaluation provides a reliable and scalable
framework for assessing multimodal generation models \cite{hua2025mmig}.
Building on this line, we introduce a multi-stage
VQA metric driven by an LLM evaluator (Gemini-3.1-Pro~\cite{gemini31pro});
an overview is illustrated in \cref{fig:vqa_pipeline}.

Given a generated clip, an LLM question generator conditions on the
prompt and first frame to produce 24 polarity-balanced Yes/No
questions (six per dimension).
To mitigate acquiescence bias, we adopt mixed polarity:
positive questions verify expected behaviors
(\textit{Yes} $\rightarrow$ Pass), while negative questions probe
failure modes (\textit{Yes} $\rightarrow$ Fail).

The question bank is refined through three stages.
\textbf{(1) Ground-truth filtering:}
the evaluator answers each question using the ground-truth video and
removes those answered incorrectly, ensuring self-consistency.
\textbf{(2) Failure filtering:}
the remaining questions are tested on curated failure clips from the
same scene, and questions that fail to penalize known errors are removed.
\textbf{(3) Human cross-validation:}
Ph.D.-level researchers and experienced AI engineers review the
refined question bank together with the failure cases to verify that
each question is unambiguous, correctly polarized, and answerable
from the video. The final validated question bank is then applied by an LLM scorer to
each generated video, producing per-dimension pass rates.

\minisection{Dimensions.}
The VQA-based evaluation covers four dimensions:

\noindent\underline{Instruction Following} 
assesses whether the generated video faithfully executes the
spatiotemporal instructions specified in the prompt, including
camera motions, subject trajectories, and ordered events.

\noindent\underline{Object \& Background Consistency}
probes the consistency of foreground objects and background elements
across frames, detecting artifacts such as morphing, identity switches,
or unexpected scene changes.

\noindent\underline{Continuity of Memory}
measures object permanence—whether the model maintains the identity,
trajectory, and state of a subject after it disappears from the
field of view and before it reappears.
This dimension most directly aligns with the
disappear-and-reappear paradigm of \NAME.

\noindent\underline{Physics Adherence}
evaluates physical plausibility, including natural locomotion,
consistent gravity, and coherent lighting and shadows as subjects
move through the scene.

\begin{figure}[t]
    \centering
    \includegraphics[width=\linewidth]{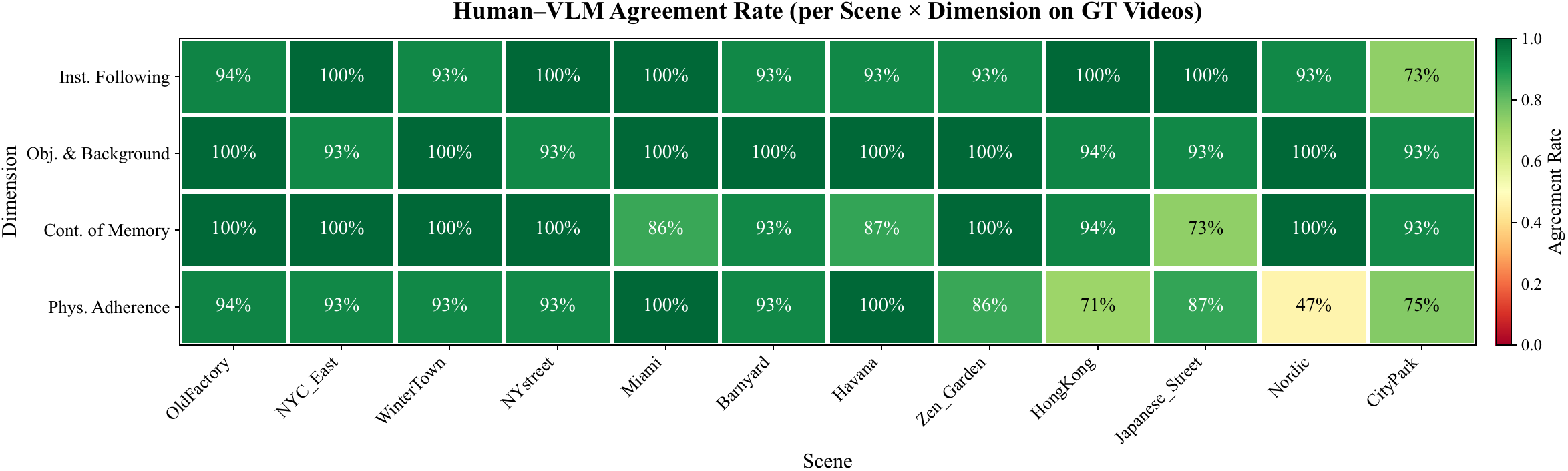}
    \caption{\textbf{Human--VLM agreement on ground-truth videos.}
Agreement rate per scene and dimension across 30 human responses.
Overall agreement reaches 92.9\%, indicating strong alignment
between our VQA-based evaluation and human judgments.}
    \label{fig:human_vlm}
\end{figure}
\minisection{Human Validation.}
To validate the reliability of our VLM-generated questions and
ground-truth answers, we conduct a human correlation study.
We randomly sample 96 questions (8 per scene across 12 scenes),
covering all four dimensions with mixed polarity, and distribute them
across four interleaved survey versions.
In total, 30 responses are collected from Ph.D.-level researchers
and experienced AI engineers, each answering Yes/No on the
ground-truth videos. \cref{fig:human_vlm} shows the per-scene, per-dimension agreement
between human majority answers and VLM-generated ground-truth answers.
The results yield an overall agreement of 92.9\% with
Cohen's $\kappa = 0.85$, confirming that our VQA evaluation
closely aligns with human judgment.

\section{Evaluation Results}

\minisection{Testing Models.}
We evaluate ten world generation models on \NAME across two categories.
We assess seven control-conditioned text-image-to-video (C+TI2V) models,
which accept explicit control signals such as camera poses or actions:
LingBot-World~\cite{lingbot-world}, Wan2.2~\cite{wan2025wan},
FantasyWorld~\cite{dai2025fantasyworld},
HunyuanWorldPlay~\cite{hyworld2025},
HunyuanGameCraft~\cite{li2025hunyuan},
the action-conditioned Matrix-Game~2.0~\cite{he2025matrix}, and
the pose-conditioned novel-view-synthesis model
Stable Virtual Camera~\cite{zhou2025stable};
and three open-source text-image-to-video (TI2V) models without explicit
control conditioning: Open-SoRA~\cite{peng2025open},
LTX-Video~\cite{hacohen2024ltx}, and CogVideoX~\cite{yang2024cogvideox}.
Implementation details and generation configurations are provided in the supplementary material (\cref{sec:model-config} and \cref{sec:implement}).

\subsection{Analysis of Automated Metrics}
\label{sec:analysis-auto-metric}

\begin{table*}[t]
\centering
\caption{
  \textbf{Automated evaluation of 10 world generation models on \NAME.}
  Models are grouped into C+TI2V and TI2V categories.
  $\uparrow$: higher is better; $\downarrow$: lower is better.
  \textbf{Bold}: best; \underline{underline}: second best.
}
\setlength{\tabcolsep}{0.45em}
\renewcommand{\arraystretch}{1.2}
\begin{adjustbox}{width=\linewidth}
\begin{tabular}{l cccc ccccc c}
\toprule
& \multicolumn{4}{c}{General Video Quality}
& \multicolumn{5}{c}{Memory-Specific}
& \multicolumn{1}{c}{Prompt} \\
\cmidrule(lr){2-5}\cmidrule(lr){6-10}\cmidrule(lr){11-11}
\textbf{Model}
  & VisQual$\,\uparrow$
  & MotSmooth$\,\uparrow$
  & ObjConsist$\,\uparrow$
  & 3DConsist$\,\uparrow$
  & ORS$\,\uparrow$
  & PSNR$\,\uparrow$
  & SSIM$\,\uparrow$
  & LPIPS$\,\downarrow$
  & CamCtrl$\,\uparrow$
  & ImgReward$\,\uparrow$ \\
\midrule

\rowcolor[HTML]{e9edf6}
\multicolumn{11}{c}{\textbf{C+TI2V Models}} \\
\midrule
LingBot-World~\cite{lingbot-world}
  & 47.4 & 57.6 & 59.0 & 88.2
  & \underline{0.381} & \underline{14.41} & \underline{0.490} & \underline{0.482} & 37.4 & \underline{36.7} \\
Wan2.2~\cite{wan2025wan}
  & 40.0 & 54.0 & 50.7 & 84.5
  & 0.328 & 13.76 & 0.469 & 0.529 & 29.8 & 26.1 \\
FantasyWorld~\cite{dai2025fantasyworld}
  & 51.0 & 55.2 & 47.6 & 88.7
  & 0.276 & 13.23 & 0.427 & 0.571 & 27.2 & 30.7 \\
HunyuanWorldPlay~\cite{hyworld2025}
  & 43.5 & 66.6 & \underline{61.5} & 90.6
  & \textbf{0.582} & 14.35 & 0.471 & 0.505 & \textbf{69.9} & 24.5 \\
HunyuanGameCraft~\cite{li2025hunyuan}
  & \underline{54.2} & 51.9 & 46.6 & 85.9
  & 0.266 & 12.81 & 0.388 & 0.603 & 54.2 & 8.9 \\
Matrix-Game~2.0~\cite{he2025matrix}
  & \textbf{61.2} & \underline{83.6} & 46.5 & 93.7
  & 0.157 & 13.49 & 0.376 & 0.550 & 17.3 & 22.3 \\
Stable Virtual Camera~\cite{zhou2025stable}
  & 43.3 & 63.1 & 59.5 & 88.5
  & 0.294 & \textbf{15.36} & \textbf{0.523} & \textbf{0.455} & \underline{65.2} & 22.3 \\

\midrule
\rowcolor[HTML]{e9edf6}
\multicolumn{11}{c}{\textbf{TI2V Models}} \\
\midrule
Open-SoRA~\cite{peng2025open}
  & 49.7 & 68.3 & 47.2 & 89.7
  & 0.182 & 12.54 & 0.384 & 0.566 & 16.8 & 31.3 \\
LTX-Video~\cite{hacohen2024ltx}
  & 44.9 & \textbf{84.4} & \textbf{81.6} & \textbf{94.1}
  & 0.330 & 13.42 & 0.455 & 0.518 & 17.1 & \textbf{37.1} \\
CogVideoX~\cite{yang2024cogvideox}
  & 40.1 & 59.8 & 54.0 & \underline{94.0}
  & 0.251 & 12.07 & 0.480 & 0.592 & 12.0 & 34.9 \\
\bottomrule
\end{tabular}
\end{adjustbox}
\label{tab:auto_metrics}
\end{table*}

\begin{figure*}[t]
  \centering
  \includegraphics[width=0.95\textwidth]{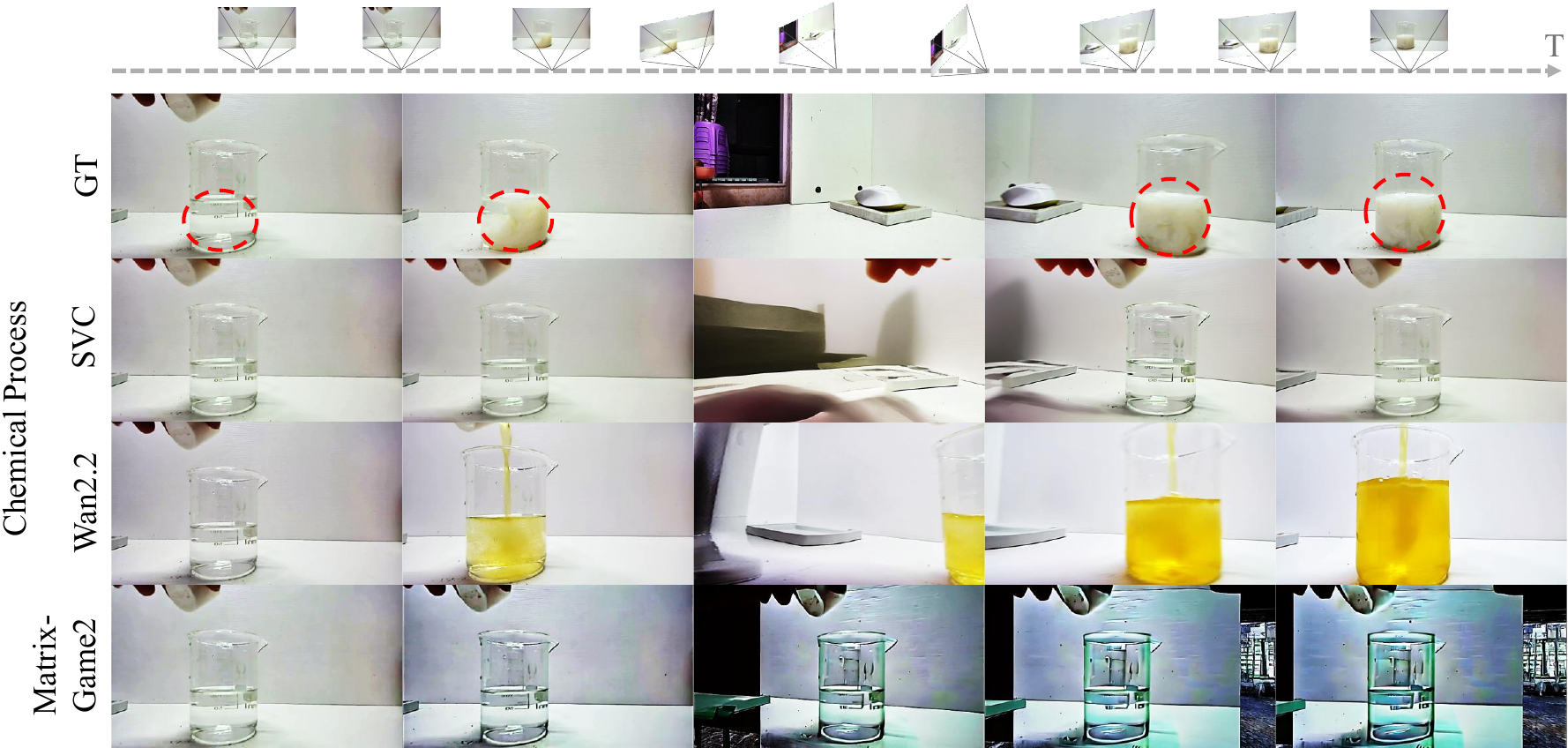}
  \caption{\textbf{Qualitative comparison of camera controllability on a real-world clip.} SVC follows the prescribed trajectory closely, while Wan2.2 and Matrix-Game~2.0 fail to reproduce the intended viewpoint changes. }
  \label{fig:qual_camctrl}
\end{figure*}
\minisection{The conditioning interface determines trajectory precision.}
View synthesis pipelines such as Stable Virtual Camera generate views directly conditioned on target camera poses, following the specified trajectory by construction. As shown in Table~\ref{tab:auto_metrics}, this leads to strong Camera Controllability and the highest pixel-level fidelity, although HunyuanWorldPlay achieves the highest overall Camera Controllability. In contrast, Matrix-Game~2.0, which receives control through action inputs rather than explicit camera poses, achieves controllability comparable to TI2V models. The key difference lies in the conditioning interface: when viewpoint changes are induced indirectly through action-conditioned dynamics rather than explicit pose conditioning, the prescribed trajectory is not translated into precise camera motion. What ultimately determines trajectory precision is whether the model's
conditioning mechanism directly exposes geometric degrees of freedom, or instead relies on implicit, learned transitions. This is visually confirmed in \cref{fig:qual_camctrl}: Stable Virtual Camera reproduces the GT pan-away-and-return trajectory with consistent viewpoint progression, whereas Matrix-Game~2.0 drifts to an different scene under the same target trajectory.

\begin{figure*}[t]
  \centering
  \includegraphics[width=0.9\textwidth]{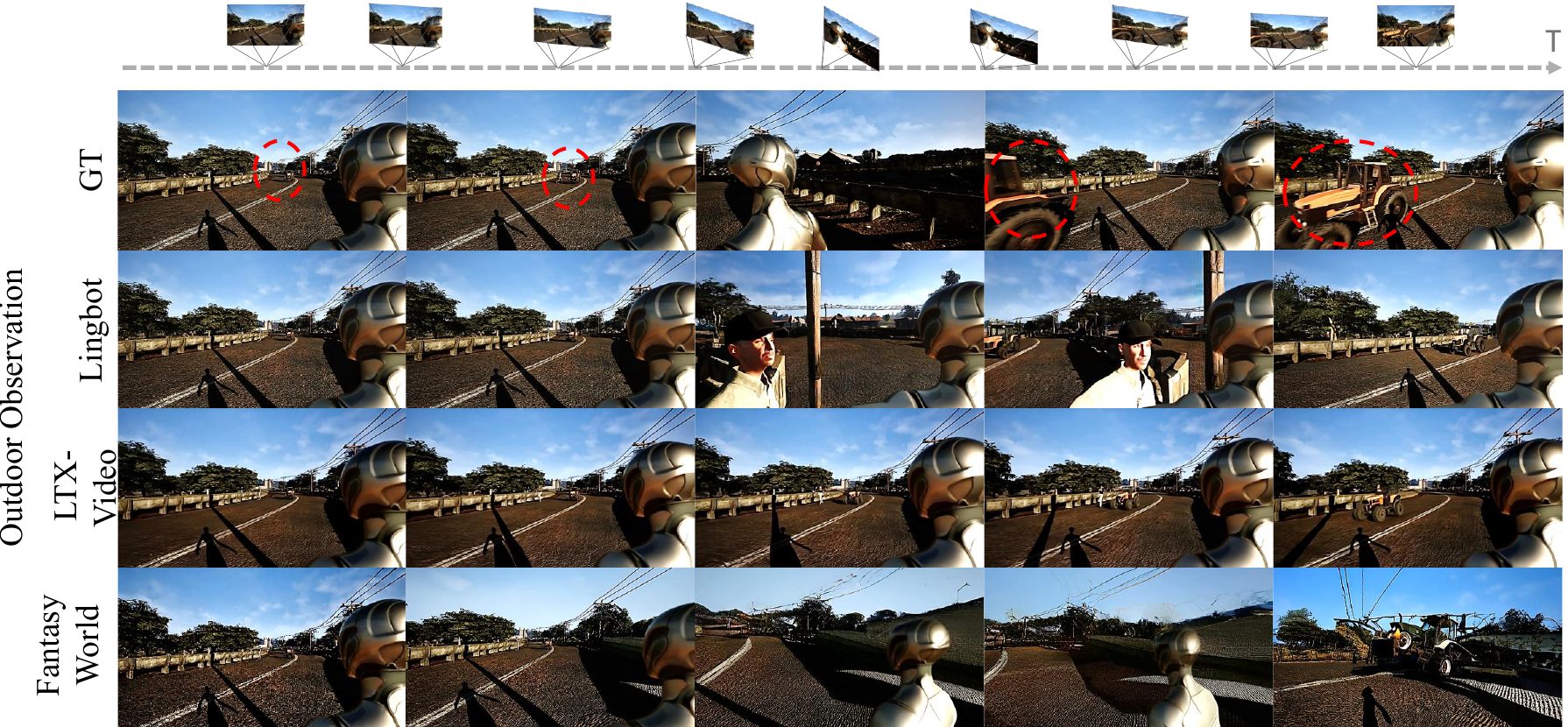}
  \caption{\textbf{Camera inactivity vs. active trajectory following.} LTX-Video produces nearly static frames, while LingBot-World and FantasyWorld follow the trajectory but fail to recover the target object upon reappearance.}
  \label{fig:qual_ors}
\end{figure*}

\minisection{Camera inactivity inflates consistency metrics.}
LTX-Video tops three of four General Video Quality metrics (Table~\ref{tab:auto_metrics}) and also obtains a relatively high ORS of 0.330, despite having Camera
Controllability comparable to the other TI2V baselines. This contradiction arises because LTX-Video barely moves the camera: when consecutive frames are nearly identical, flow-based smoothness, depth consistency, and identity similarity are trivially maximized. The same mechanism inflates its ORS, since the target object never leaves the frame and SAM-3 detects it throughout the R phase by default. This exposes a limitation of standard video quality metrics: they cannot distinguish a model that preserves appearance across genuine viewpoint changes from one that simply avoids moving. \cref{fig:qual_ors} illustrates this behavior, with LTX-Video retaining
a nearly fixed viewpoint throughout the sequence.

\begin{figure*}[t]
  \centering
  \includegraphics[width=0.9\textwidth]{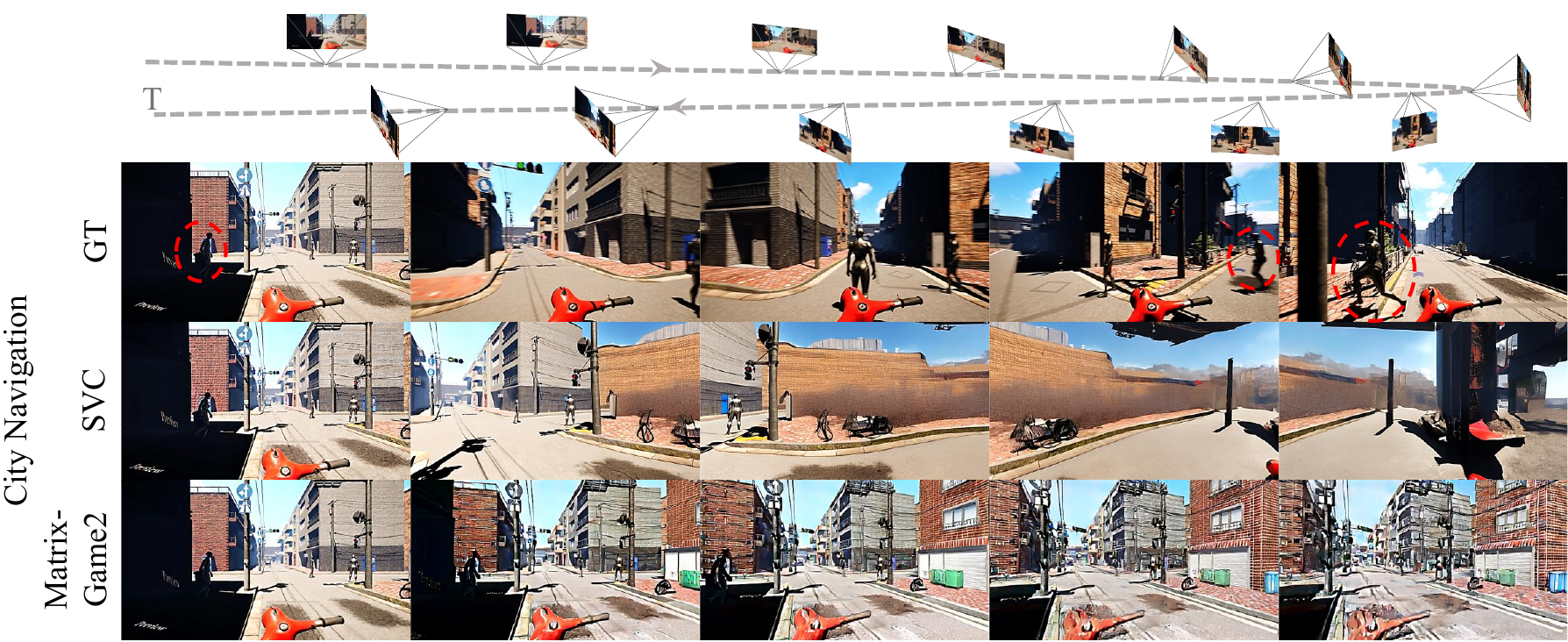}
  \caption{\textbf{Qualitative comparison of geometric fidelity and perceptual quality on a synthetic clip.} SVC preserves scene geometry but introduces artifacts, while Matrix-Game~2.0 produces visually sharper frames but drifts from the GT viewpoint.}
  \label{fig:qual_tradeoff}
\end{figure*}

\minisection{A trade-off emerges between geometric fidelity and perceptual quality.}
Stable Virtual Camera leads all pixel-level metrics, yet its Visual Quality score remains relatively low in Table~\ref{tab:auto_metrics}, whereas Matrix-Game~2.0 achieves the highest Visual Quality but ranks among the lowest in SSIM. Both models also obtain relatively low ImageReward scores. This pattern suggests that geometric consistency and perceptual naturalness are not yet aligned in current methods. HunyuanGameCraft further demonstrates this metric mismatch: it achieves the second-highest Visual Quality score, while obtaining the lowest ImageReward and the weakest GT-aligned pixel fidelity among the camera-conditioned C+TI2V models. These results show that no-reference visual quality, prompt-image alignment, and GT-aligned geometric fidelity capture different aspects of generation quality and should not be interpreted interchangeably. As illustrated in \cref{fig:qual_tradeoff}, Matrix-Game~2.0 produces sharp frames but drifts from the GT viewpoint, whereas Stable Virtual Camera better preserves scene geometry while introducing rendering artifacts such as blurring, seams, and depth-inpainting errors.

\minisection{Camera conditioning alone does not ensure object memory.}
Five of the C+TI2V models (LingBot-World, Wan2.2, FantasyWorld, HunyuanWorldPlay, and HunyuanGameCraft) share explicit camera conditioning, yet their object memory performance varies notably
(Table~\ref{tab:auto_metrics}): HunyuanWorldPlay achieves the highest ORS among them, while LingBot-World leads all their pixel-level fidelity metrics. In contrast, FantasyWorld achieves higher Visual Quality than LingBot-World but a substantially lower ORS. This gap within the same model category reveals that camera conditioning does not by itself encourage the model to maintain a representation of objects that have left the field of view. A model can produce aesthetically better frames while failing to recall what it previously observed, suggesting that object permanence must be explicitly targeted during training rather than as a byproduct of camera-conditioned generation. As shown in \cref{fig:qual_ors}, both LingBot-World and FantasyWorld receive the same camera trajectory, yet LingBot-World produces a recognizable return to the target region while FantasyWorld generates frames that bear little resemblance to the ground-truth reappearance.

\minisection{ORS reveals memory failures and reliable reappearance remains open.}
No model exceeds an ORS of 0.6 (Table~\ref{tab:auto_metrics}),
indicating that even the top performer does not reliably re-detect the target object throughout the R phase. However, ORS must be interpreted jointly with Camera Controllability: LTX-Video obtains an ORS of 0.330 despite its low Camera Controllability, indicating that part of its score can be attributed to camera inactivity rather than genuine disappearance and reappearance. Among models that actually execute the trajectory, HunyuanWorldPlay leads, followed by LingBot-World, Wan2.2, and Stable Virtual Camera. The low absolute values indicate that current models lack a persistent internal representation of disappeared objects. Once the target leaves the frame, the model's ``memory'' degrades rapidly, and the reappeared content is either absent, hallucinated, or unrecognizable. As shown in \cref{fig:qual_ors}, even LingBot-World, one of the top-performing models on ORS, fails to recover the target object faithfully upon reappearance, and LTX-Video's apparent success reflects a static viewpoint rather than genuine object recall. Overall, no single model simultaneously achieves strong Camera Controllability, high ORS, and competitive Visual Quality. Closing this gap is a core challenge that \NAME exposes for future world generation models.

\subsection{Analysis of VQA-based Evaluation}
\label{sec:analysis-vqa-metric}

\minisection{Camera inactivity inflates VQA scores; Instruction Following exposes the gap.}
LTX-Video achieves the highest scores on two of the four VQA
dimensions (Table~\ref{tab:vqa}) and ranks second on Physics
Adherence, narrowly behind HunyuanWorldPlay. As a TI2V model without camera conditioning, its strong consistency-oriented scores mirror the inflation pattern observed in automated metrics. However, Instruction Following reveals a different ranking: LingBot-World leads, closely followed by HunyuanWorldPlay, and C+TI2V models occupy the top four positions, while TI2V models cluster at lower scores. This indicates that camera conditioning generally improves the execution of spatiotemporal instructions, whereas consistency-oriented VQA scores can be inflated when a model avoids the requested viewpoint change. Notably, LingBot-World's advantage in Instruction Following does not transfer to other dimensions: its Object \& Background score remains substantially lower than that of LTX-Video, suggesting that actively following the trajectory introduces inconsistencies that static models avoid by not moving.
\begin{table*}[htbp]
\centering
\caption{%
  \textbf{VQA evaluation across four semantic dimensions on \NAME.}
  Each dimension is scored 0--100 ($\uparrow$: higher is better).
  Models are grouped into C+TI2V and TI2V categories.
  \textbf{Bold}: best; \underline{underline}: second best.
}
\resizebox{0.75\textwidth}{!}{%
\begin{tabular}{l cccc}
\toprule
\textbf{Model}
  & Inst.Fol.$\,\uparrow$
  & Obj.\&Bkg.$\,\uparrow$
  & Cont.Mem.$\,\uparrow$
  & Phys.Adh.$\,\uparrow$ \\
\midrule

\rowcolor[HTML]{e9edf6}
\multicolumn{5}{c}{\textbf{C+TI2V Models}} \\
\midrule
LingBot-World~\cite{lingbot-world}   & \textbf{64.2} & 44.4 & 42.1 & 53.6 \\
Wan2.2~\cite{wan2025wan}             & 50.6 & 30.2 & 36.8 & 38.9 \\
FantasyWorld~\cite{dai2025fantasyworld} & 50.5 & 25.6 & 37.1 & 33.6 \\
HunyuanWorldPlay~\cite{hyworld2025} & \underline{61.6} & 66.4 & \underline{55.6} & \textbf{63.6} \\
HunyuanGameCraft~\cite{li2025hunyuan} & 41.6 & \underline{71.6} & 48.4 & 61.0 \\
Matrix-Game~2.0~\cite{he2025matrix}  & 37.5 & 12.7 & 36.5 & 21.8 \\
Stable Virtual Camera~\cite{zhou2025stable} & 49.7 & 23.8 & 29.6 & 33.3 \\
\midrule

\rowcolor[HTML]{e9edf6}
\multicolumn{5}{c}{\textbf{TI2V Models}} \\
\midrule
Open-SoRA~\cite{peng2025open}        & 43.2 & 66.8 & 48.3 & 59.7 \\
LTX-Video~\cite{hacohen2024ltx}      & 41.0 & \textbf{76.6} & \textbf{57.0} & \underline{63.5} \\
CogVideoX~\cite{yang2024cogvideox}   & 40.5 & 52.4 & 42.7 & 42.8 \\
\bottomrule
\end{tabular}%
}
\label{tab:vqa}
\end{table*}

\minisection{Semantic evaluation reveals artifacts missed by automated metrics.}
Matrix-Game~2.0 records the lowest Object \& Background and
Physics Adherence scores across all models (\cref{tab:vqa}),
despite achieving the highest Visual Quality and second-highest
Motion Smoothness in automated evaluation.
Stable Virtual Camera shows a similar trend: strong pixel-level
fidelity but below-average VQA scores.
These results suggest that rendering artifacts—such as warping seams,
depth inpainting errors, and texture flickering—are largely invisible
to no-reference quality metrics but are penalized by VQA evaluation,
which focuses on semantic correctness rather than perceptual sharpness.

\minisection{Continuity of Memory remains a major bottleneck.}
The highest Continuity of Memory score is achieved by LTX-Video,
whose score may be inflated by camera inactivity
(\cref{tab:vqa}). Among models that actively follow the trajectory, HunyuanWorldPlay achieves the highest score, although only slightly more than half of the memory-related questions are answered correctly. Together with the low ORS values observed in automated evaluation, this result confirms that current models fail to maintain a reliable representation of objects once they leave the field of view, both at the signal level and the semantic level.
\section{Conclusion}
This paper introduces \NAME, a novel benchmark that evaluates memory consistency
in world generation through the disappear-and-reappear paradigm.
By combining automated metrics with a VQA pipeline across 360
clips, we found that current models struggle to maintain a
persistent representation of objects that leave the field of view.
No model exceeds an Object Reappearance Score of 0.6, and models
without camera conditioning inflate consistency scores by
generating near-static video rather than executing viewpoint
changes. Even among camera-conditioned models, object permanence
does not emerge as a byproduct of trajectory control, indicating
that memory must be explicitly addressed in model design. 
These findings point to future work on persistent state representations, 
memory-aware training objectives, and evaluation protocols that account for camera inactivity.
We hope \NAME serves as a useful tool for tracking progress
on these challenges.
\section*{Acknowledgements}
    YZ was supported in part by the SoftBank Group–ARM Fellowship.
    This work was supported in part by the Office of Naval Research (ONR) under Grant No.~N000142412696 and also by a gift from the Chan Zuckerberg Initiative Foundation to establish the Kempner Institute for the Study of Natural and Artificial Intelligence at Harvard University.

\clearpage


%
%
\bibliographystyle{splncs04}
\bibliography{main}

\clearpage
\renewcommand{\thepage}{S\arabic{page}}
\setcounter{page}{1}

\setcounter{section}{0}
\renewcommand{\thesection}{\Alph{section}}
\renewcommand{\theHsection}{supp.\Alph{section}}
\renewcommand{\theHsubsection}{supp.\Alph{section}.\arabic{subsection}}
\renewcommand{\theHfigure}{supp.S\arabic{figure}}
\renewcommand{\theHtable}{supp.S\arabic{table}}

\renewcommand{\thefigure}{S\arabic{figure}}
\renewcommand{\thetable}{S\arabic{table}}
\setcounter{table}{0}
\setcounter{figure}{0}
\raggedbottom
\section{Supplementary Materials}

\vspace{0.5em}
\noindent\textbf{Table of Contents}
\vspace{0.5em}

\newcommand{\tocline}[3]{\noindent\hyperref[#3]{\makebox[3.2em][l]{#1}#2}\dotfill\pageref{#3}\\}
\newcommand{\tocsubline}[3]{\noindent\hspace{2em}\hyperref[#3]{\makebox[3.8em][l]{#1}#2}\dotfill\pageref{#3}\\}

{\small
\tocline{A.1}{Model Configurations}{sec:model-config}
\tocline{A.2}{Implementation Details}{sec:implement}
\tocline{A.3}{Dataset Statistics}{sec:supp-dataset}
\tocline{A.4}{More Dataset Examples}{sec:more_dataset}
\tocline{A.5}{More Qualitative Results}{sec:more-qualitative}
\tocline{A.6}{Additional Radar Visualizations}{sec:supp-radar}
\tocline{A.7}{Ablation Studies}{sec:ablation}
\indent \tocsubline{A.7.1}{ORS Robustness Analysis}{sec:ablation_ors}
\indent \tocsubline{A.7.2}{Motion-Gated Evaluation}{sec:ablation_motion}
\indent \tocsubline{A.7.3}{Per-Phase Fidelity Breakdown}{sec:ablation_phase}
\indent \tocsubline{A.7.4}{Metric Sensitivity Analysis}{sec:ablation_metric}
\indent \tocsubline{A.7.5}{Camera Pose Estimation Validation}{sec:ablation_pose}
\indent \tocsubline{A.7.6}{Initial-State Conditioning vs. Backbone Capacity}{sec:ablation_capacity}
\tocline{A.8}{Detailed VQA Pipeline}{sec:complete-vqa}
\tocline{A.9}{Failure Analysis}{sec:failure-analysis}
}
\clearpage

\subsection{Model Configurations}
\label{sec:model-config}

We evaluate ten open-source models spanning two categories: control-conditioned text-image-to-video generation (C+TI2V) and standard text-image-to-video generation (TI2V). Table~\ref{tab:model_config} summarizes the key specifications of each model, including output resolution, frame rate, generated video length, and whether the model supports explicit camera pose conditioning.

Within the C+TI2V category, the five camera-conditioned video models (LingBot-World, Wan2.2, FantasyWorld, HunyuanWorldPlay, and HunyuanGameCraft) accept camera trajectories as input, allowing direct control over viewpoint changes. Stable Virtual Camera is likewise pose-conditioned, but is trained for static-scene novel view synthesis (NVS) without text conditioning or temporal dynamics, while Matrix-Game~2.0 receives control through action inputs rather than explicit camera poses. The TI2V models (Open-SoRA, LTX-Video, and CogVideoX) do not support camera conditioning and instead rely on text prompts or learned priors to determine camera motion. Including these non-camera-conditioned baselines allows us to assess how much explicit camera control contributes to generation quality and geometric consistency.

\begin{table}[h]
\centering
\caption{
  \textbf{Model configurations used in our evaluation.}
  We summarize the output resolution, frame rate, video length, and camera-conditioning support for each baseline. All models are open-source. C+TI2V and TI2V denote control-conditioned and standard text-image-to-video generation, respectively.
}
\setlength{\tabcolsep}{6pt}
\renewcommand{\arraystretch}{1.3}
\begin{adjustbox}{width=\linewidth}
\begin{tabular}{l l r r l c c}
\toprule
\textbf{Model} & \textbf{Ability}
  & \textbf{Res.} & \textbf{FPS}
  & \begin{tabular}[c]{@{}c@{}}\textbf{Length}\\\textbf{(s\,/\,frames)}\end{tabular}
  & \begin{tabular}[c]{@{}c@{}}\textbf{Open}\\\textbf{Source}\end{tabular}
  & \textbf{Camera} \\
\midrule
LingBot-World~\cite{lingbot-world} & C+TI2V & $464\times832$ & 16 & 5.1\,/\,81  & \textcolor{green!70!black}{\checkmark} & \textcolor{green!70!black}{\checkmark} \\
Wan2.2~\cite{wan2025wan}        & C+TI2V & $480\times832$ & 16 & 5.1\,/\,81  & \textcolor{green!70!black}{\checkmark} & \textcolor{green!70!black}{\checkmark} \\
FantasyWorld~\cite{dai2025fantasyworld}  & C+TI2V & $480\times832$ & 16 & 5.1\,/\,81  & \textcolor{green!70!black}{\checkmark} & \textcolor{green!70!black}{\checkmark} \\
HunyuanWorldPlay~\cite{hyworld2025} & C+TI2V & $480\times848$ & 24 & 5.2\,/\,125 & \textcolor{green!70!black}{\checkmark} & \textcolor{green!70!black}{\checkmark} \\
HunyuanGameCraft~\cite{li2025hunyuan} & C+TI2V & $720\times1280$ & 24 & 4.1\,/\,99 & \textcolor{green!70!black}{\checkmark} & \textcolor{green!70!black}{\checkmark} \\
Matrix-Game~2.0~\cite{he2025matrix} & C+TI2V & $352\times640$ & 60 & 4.8\,/\,285 & \textcolor{green!70!black}{\checkmark} & \textcolor{green!70!black}{\checkmark} \\
Stable Virtual Camera~\cite{zhou2025stable} & C+TI2V & $576\times1024$ & 16 & 5.0\,/\,80 & \textcolor{green!70!black}{\checkmark} & \textcolor{green!70!black}{\checkmark} \\
Open-SoRA~\cite{peng2025open}     & TI2V & $192\times336$ & 24 & 5.4\,/\,129 & \textcolor{green!70!black}{\checkmark} & \textcolor{red!80!black}{\texttimes} \\
LTX-Video~\cite{hacohen2024ltx}     & TI2V & $480\times832$ & 25 & 5.2\,/\,129 & \textcolor{green!70!black}{\checkmark} & \textcolor{red!80!black}{\texttimes} \\
CogVideoX~\cite{yang2024cogvideox}     & TI2V & $480\times832$ & 16 & 5.1\,/\,81  & \textcolor{green!70!black}{\checkmark} & \textcolor{red!80!black}{\texttimes} \\
\bottomrule
\end{tabular}
\end{adjustbox}
\label{tab:model_config}
\end{table}

\subsection{Implementation Details}
\label{sec:implement}

All experiments are conducted on a server equipped with four NVIDIA RTX Pro 6000 GPUs.
We reproduce each baseline using its official codebase and publicly available checkpoints.
Below we summarize the inference configuration for each model.

\noindent\textbf{LingBot}~\cite{lingbot-world}.
We use the official \texttt{lingbot-world-base-cam} checkpoint and generate 81 frames at $464 \times 832$ resolution with the UniPC solver, 70 sampling steps, a guidance scale of 5.0, and 16\,fps output.

\noindent\textbf{FantasyWorld}~\cite{dai2025fantasyworld}.
We use the Wan2.1-I2V-14B-480P variant and generate 81 frames at $336 \times 592$ resolution with 50 flow-matching steps and a guidance scale of 5.0.

\noindent\textbf{HunyuanWorldPlay}~\cite{hyworld2025}.
We use the official \texttt{HY-WorldPlay} distilled checkpoint (\texttt{ar\_distilled\_action\_model})
and generate 125 frames at $480 \times 848$ resolution with 4 flow-matching steps,
a guidance scale of 1.0, and 24\,fps output.

\noindent\textbf{HunyuanGameCraft}~\cite{li2025hunyuan}.
We use the official \texttt{Hunyuan-GameCraft-1.0} distilled checkpoint (\texttt{mp\_rank\_00\_model\_states\_distill.pt})
and generate 99 frames at $720 \times 1280$ resolution with 8 flow-matching steps,
a guidance scale of 1.0, and 24\,fps output.

\noindent\textbf{Stable Virtual Camera}~\cite{zhou2025stable}.
We use the v1.1 checkpoint and run two-pass Euler EDM sampling with 50 steps each, guidance scales of 3.0 and 2.0, outputting 80 frames at $576 \times 1024$ and 16\,fps.

\noindent\textbf{Matrix-Game}~\cite{he2025matrix}.
We use the Universal distilled checkpoint with 3 denoising steps, generating videos at $352 \times 640$ and 60\,fps with duration matched to the ground-truth sequence.

\noindent\textbf{VideoX-Fun}.
We use the Wan2.2-A14B camera-control checkpoint and generate 81 frames at $480 \times 832$ with the Flow-UniPC sampler, 50 steps, and a guidance scale of 6.0.

\noindent\textbf{Open-Sora}~\cite{peng2025open}.
We use the v2.0 checkpoint and generate 129 frames at 256px (16:9) resolution with 50 rectified-flow steps, text guidance scale 7.5, image guidance scale 3.0, and 24\,fps output.

\noindent\textbf{LTX-Video}~\cite{hacohen2024ltx}.
We use the 13B (v0.9.8-dev) checkpoint and generate 129 frames at $480 \times 832$ with a two-pass multi-scale pipeline (30 + 13 steps) and dynamic guidance scales, at 25\,fps.

\noindent\textbf{CogVideoX}~\cite{yang2024cogvideox}.
We use the CogVideoX1.5-5B-I2V checkpoint and generate 81 frames at $480 \times 832$ with a DPM scheduler, 50 steps, a guidance scale of 6.0, and 16\,fps.

\subsection{Dataset Statistics}
\label{sec:supp-dataset}

\begin{figure}[h]
  \centering
  \begin{subfigure}[b]{0.455\linewidth}
    \includegraphics[width=\linewidth]{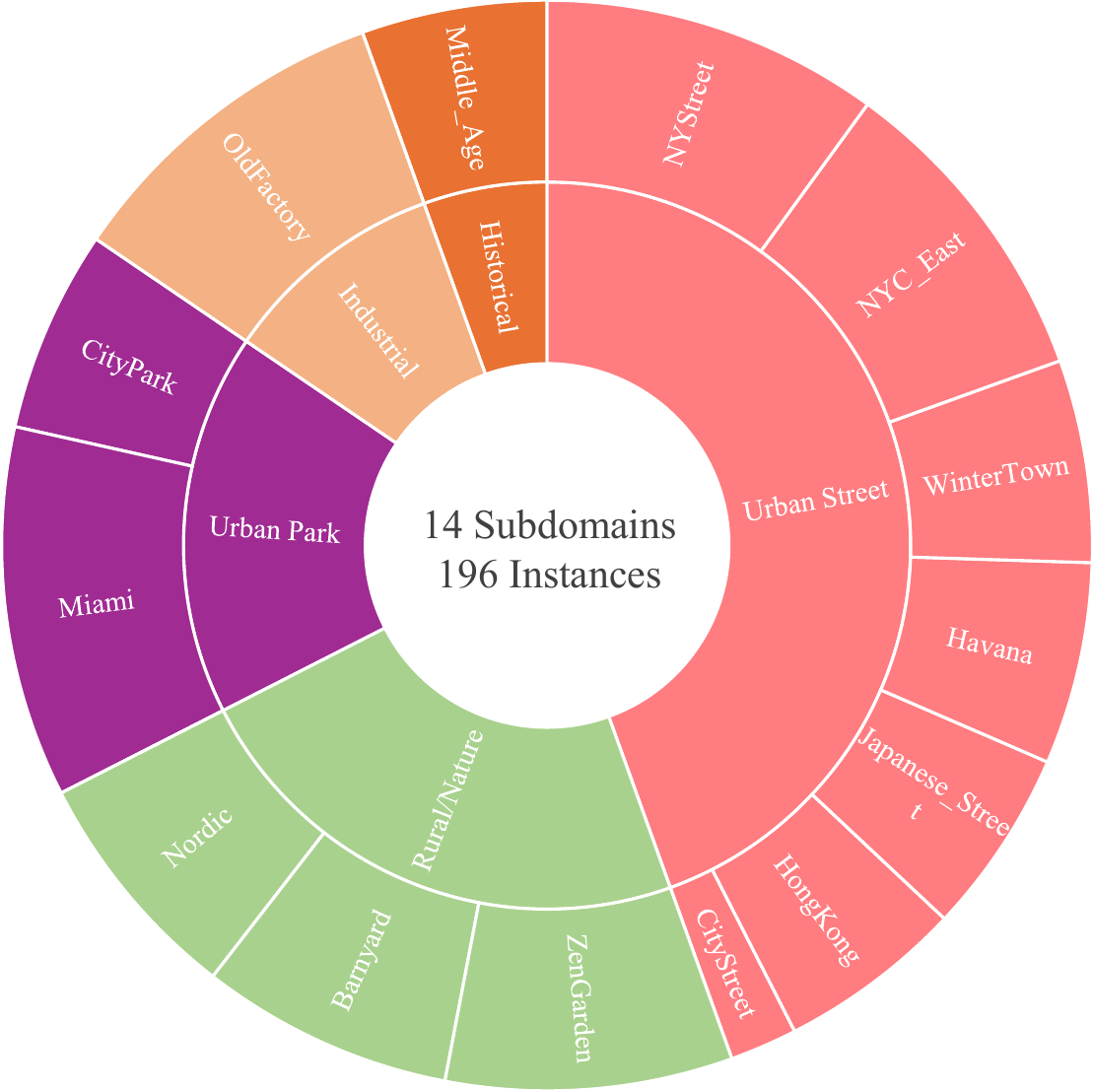}
    \caption{Synthetic: 14 scene subdomains across five environment categories.}
    \label{fig:synthetic_scenes}
  \end{subfigure}
  \hfill
  \begin{subfigure}[b]{0.455\linewidth}
    \centering
    \includegraphics[width=\linewidth]{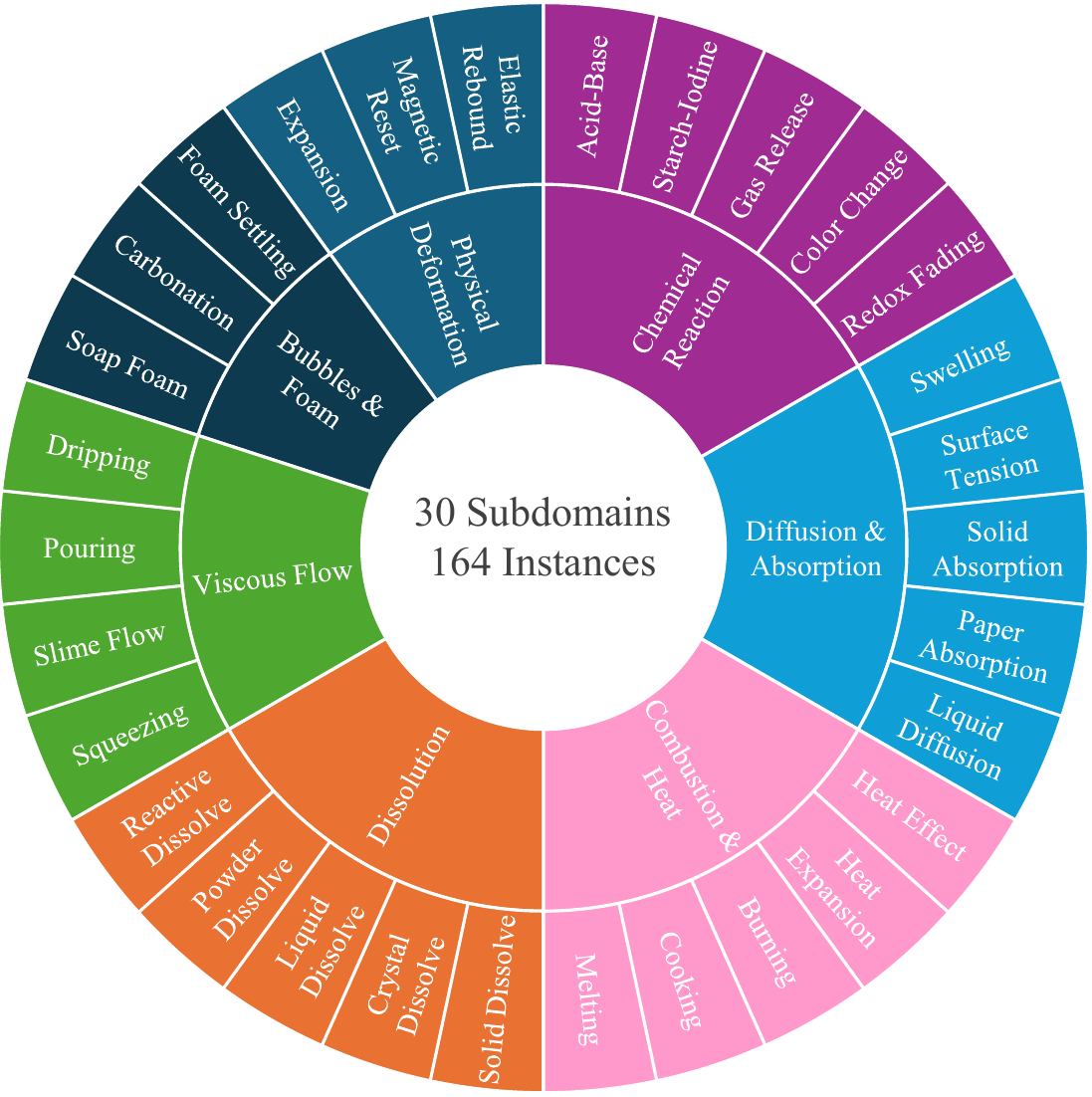}
    \caption{Real-world: 30 state-change processes across seven major categories.}
    \label{fig:real_data}
  \end{subfigure}
  \caption{Dataset overview of \NAME.}
  \label{fig:dataset_overview}
\end{figure}

The synthetic subset contains 196 clips spanning 14 scene subdomains
(\cref{fig:synthetic_scenes}) across five environment categories,
featuring diverse target objects and action types. Sequences are
typically 260--300 frames long, rendered at 1920$\times$1080 (60\,FPS).

The real-world subset contains 164 clips captured at
1920$\times$1080, emphasizing diversity of physical-state changes:
30 common state-change processes grouped into seven major categories
(\cref{fig:real_data}). Sequences range from 103--349 frames.

\noindent\textbf{Camera trajectory statistics.}
Table~\ref{tab:traj_stats} summarizes the camera-trajectory
distribution of the two subsets. The real-world subset primarily
contains controlled horizontal pans and vertical tilts, whereas the
synthetic subset contains more diverse viewpoint changes, including
U-turns, forward motion, head turns, and vertical motion. For each
trajectory type, we report the number of clips, the mean total camera
rotation, and the mean temporal gap between departure and
reappearance. The trajectory counts sum to 164 real-world clips and
196 synthetic clips, matching the sizes of the two subsets.

\begin{table}[t]
\centering
\vspace{-1mm}
\caption{
\textbf{Camera trajectory statistics of \NAME.}
Rotation denotes the mean total camera rotation, and gap denotes the
mean number of frames between departure and reappearance.
}
\label{tab:traj_stats}
\scriptsize
\setlength{\tabcolsep}{4pt}
\renewcommand{\arraystretch}{1.1}
\begin{adjustbox}{width=\linewidth}
\begin{tabular}{llrrr}
\toprule
\textbf{Subset}
& \textbf{Trajectory}
& \textbf{\# Clips}
& \textbf{Rotation ($^\circ$)}
& \textbf{Gap (frames)} \\
\midrule
Real-world
& Pan L$\rightarrow$R
& 56 & 72 & 59 \\
Real-world
& Pan R$\rightarrow$L
& 51 & 63 & 65 \\
Real-world
& Tilt U$\rightarrow$D
& 57 & 38 & 70 \\
\midrule
Synthetic
& U-turn
& 68 & 178 & 113 \\
Synthetic
& Forward
& 58 & 92 & 81 \\
Synthetic
& Head turn
& 41 & 70 & 76 \\
Synthetic
& Vertical
& 29 & 27 & 77 \\
\bottomrule
\end{tabular}
\end{adjustbox}
\end{table}

The synthetic trajectories generally involve larger viewpoint changes
and longer temporal gaps. In particular, U-turn sequences exhibit the
largest mean camera rotation ($178^\circ$) and the longest mean
departure-to-reappearance gap (113 frames), providing a challenging
setting for evaluating memory across substantial viewpoint changes.

\FloatBarrier

\subsection{More Dataset Examples}
\label{sec:more_dataset}

We provide additional dataset examples for both the synthetic and real-world subsets of \NAME.

\begin{figure*}[t]
    \centering
    \includegraphics[width=\linewidth]{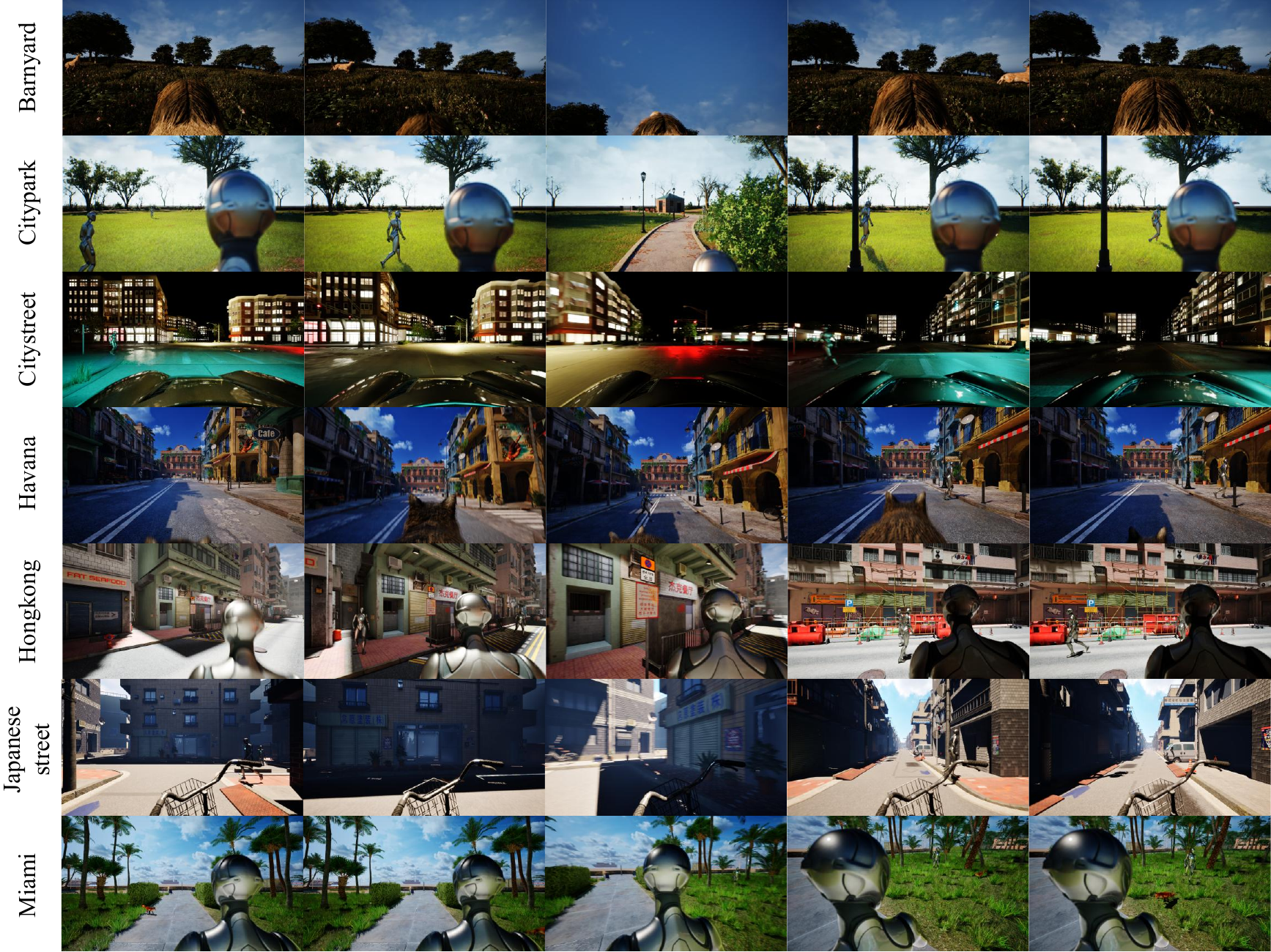}
    \caption{\textbf{Synthetic dataset examples (1/2).} Representative scene from the
    synthetic subset of \NAME, showing sampled frames across the V-D-R phases.}
    \label{fig:sup_synthetic1}
\end{figure*}

\begin{figure*}[t]
    \centering
    \includegraphics[width=\linewidth]{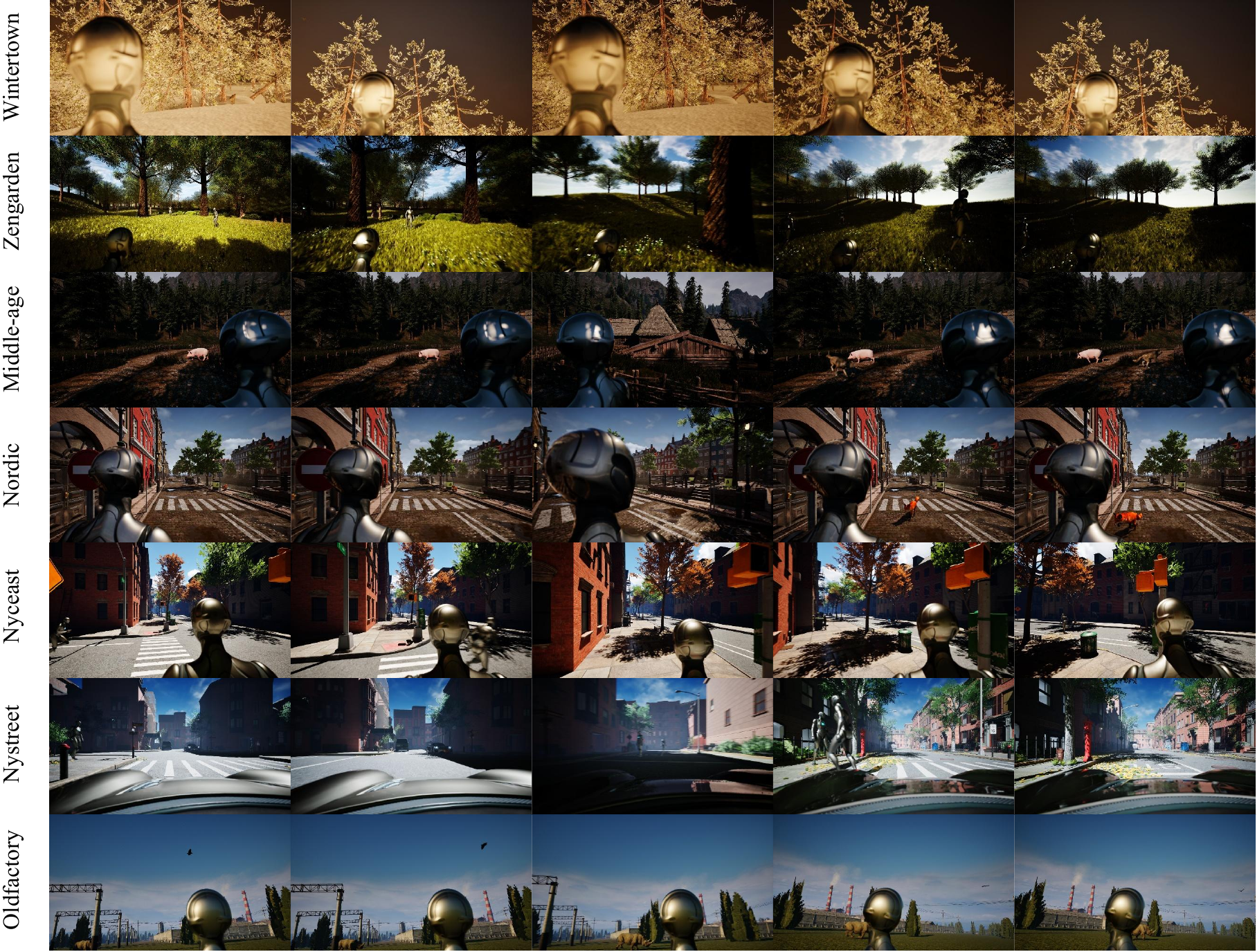}
    \caption{\textbf{Synthetic dataset examples (2/2).} Representative scene from the
    synthetic subset of \NAME, showing sampled frames across the V-D-R phases.}
    \label{fig:sup_synthetic2}
\end{figure*}

\minisection{Synthetic Data.}
\cref{fig:sup_synthetic1,fig:sup_synthetic2} show representative sequences from the synthetic scenes.
Each row displays five uniformly sampled frames from a single clip, covering the
V-phase (target object visible), D-phase (camera departs), and R-phase (camera returns).

\begin{figure*}[t]
    \centering
    \includegraphics[width=\linewidth]{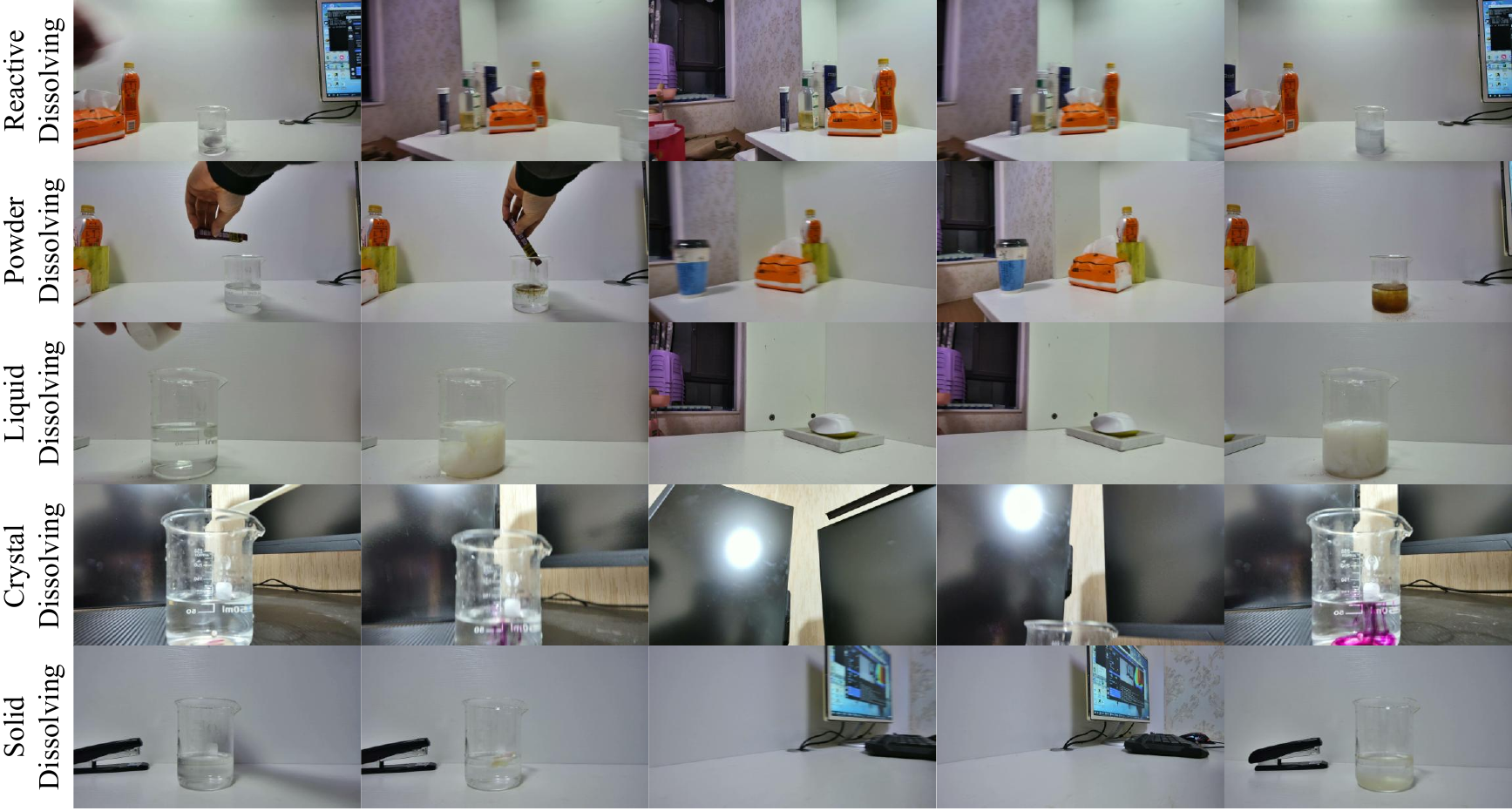}
    \caption{\textbf{Real-world examples: Dissolution.} Sampled V-D-R frames capturing dissolution processes such as salt dissolving or sugar melting, where the target object gradually loses its solid form during the camera's absence.}
    \label{fig:sup_real1}
\end{figure*}

\begin{figure*}[t]
    \centering
    \includegraphics[width=\linewidth]{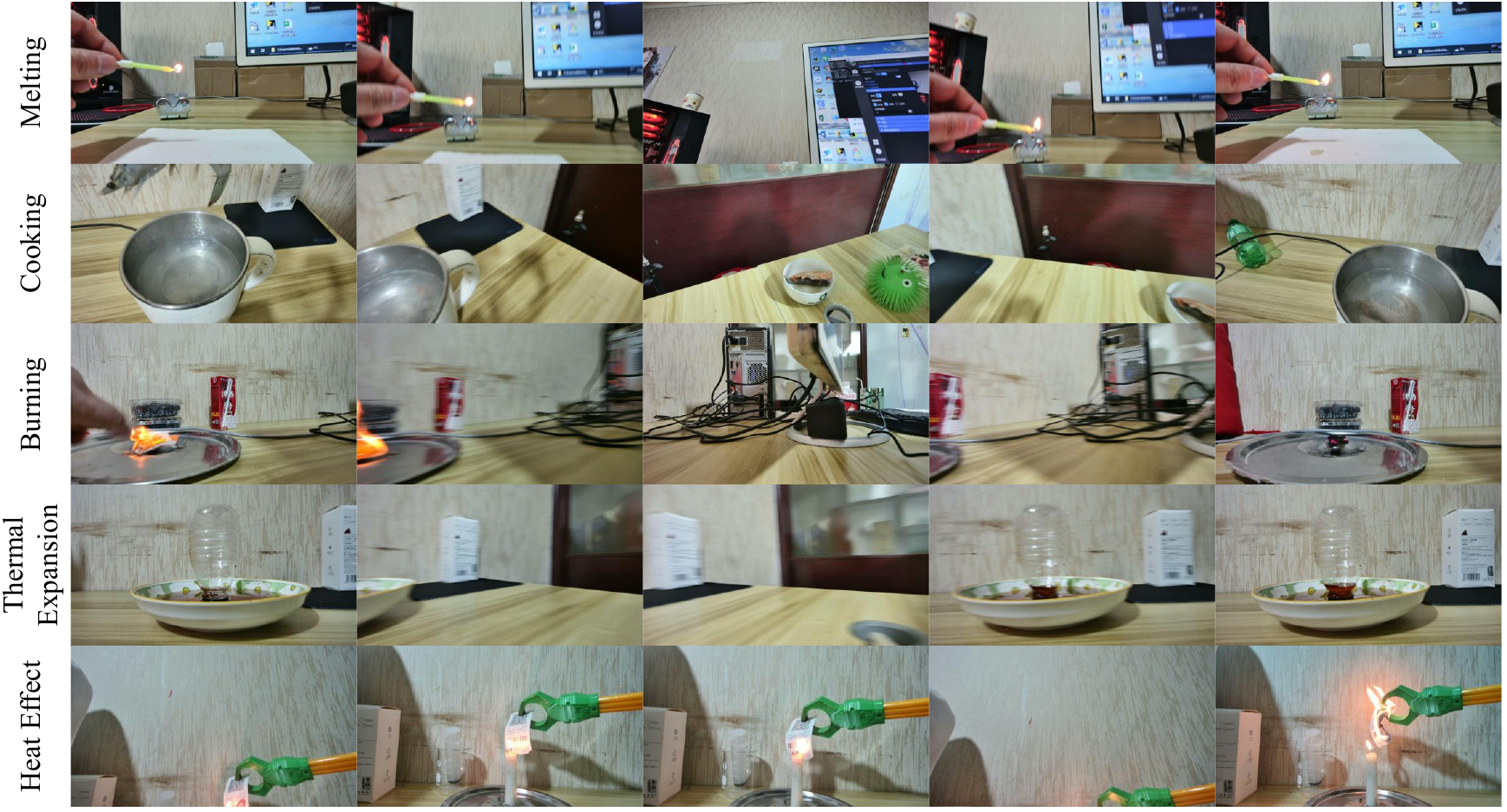}
    \caption{\textbf{Real-world examples: Combustion \& Heat.} Sampled V-D-R frames showing heat-driven state changes such as candle burning or paper burning, where the object's shape and material properties transform irreversibly.}
    \label{fig:sup_real2}
\end{figure*}

\begin{figure*}[t]
    \centering
    \includegraphics[width=\linewidth]{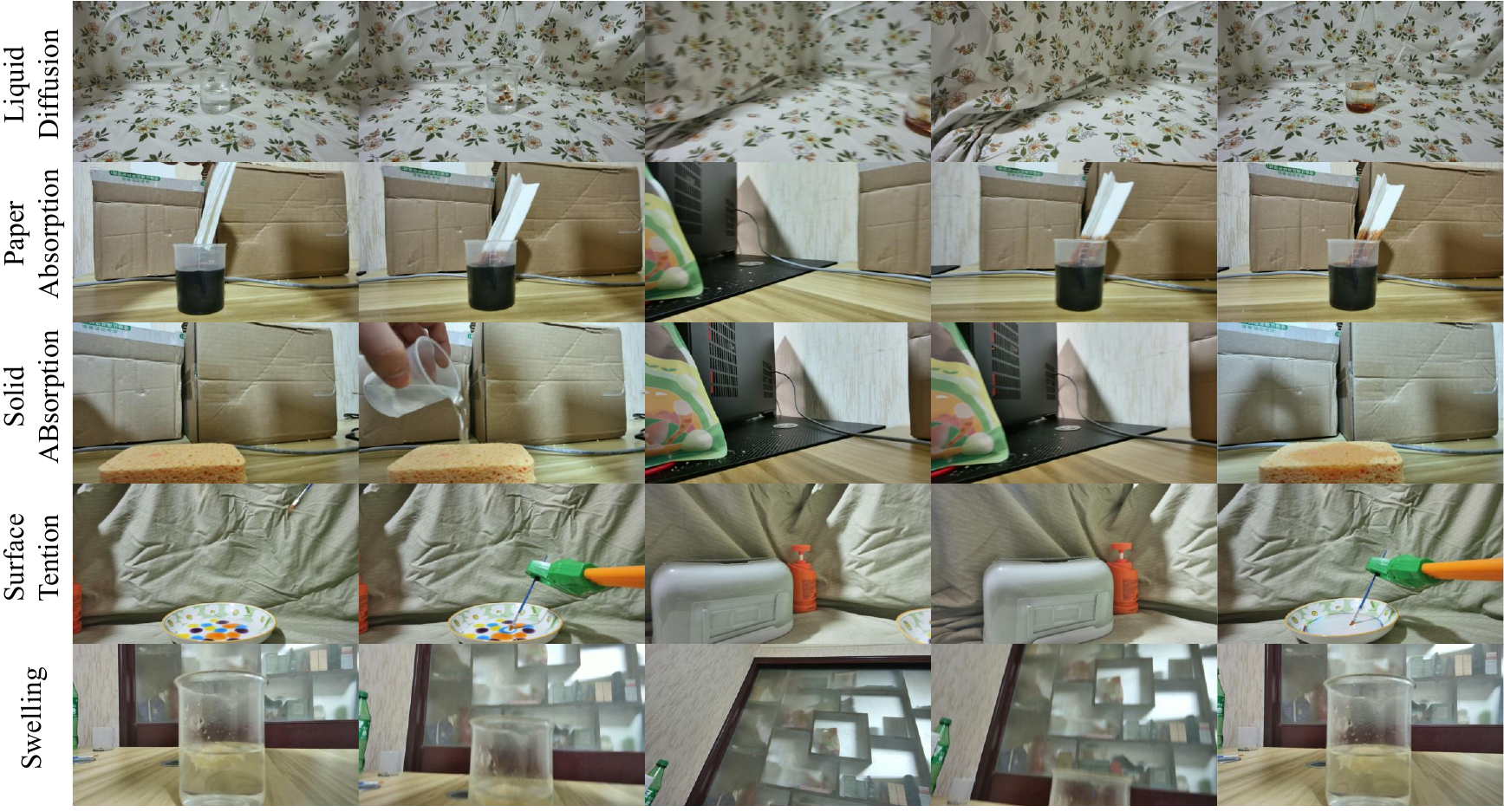}
    \caption{\textbf{Real-world examples: Diffusion \& Absorption.} Sampled V-D-R frames depicting diffusion and absorption processes such as ink spreading in water or liquid soaking into fabric.}
    \label{fig:sup_real3}
\end{figure*}

\begin{figure*}[t]
    \centering
    \includegraphics[width=\linewidth]{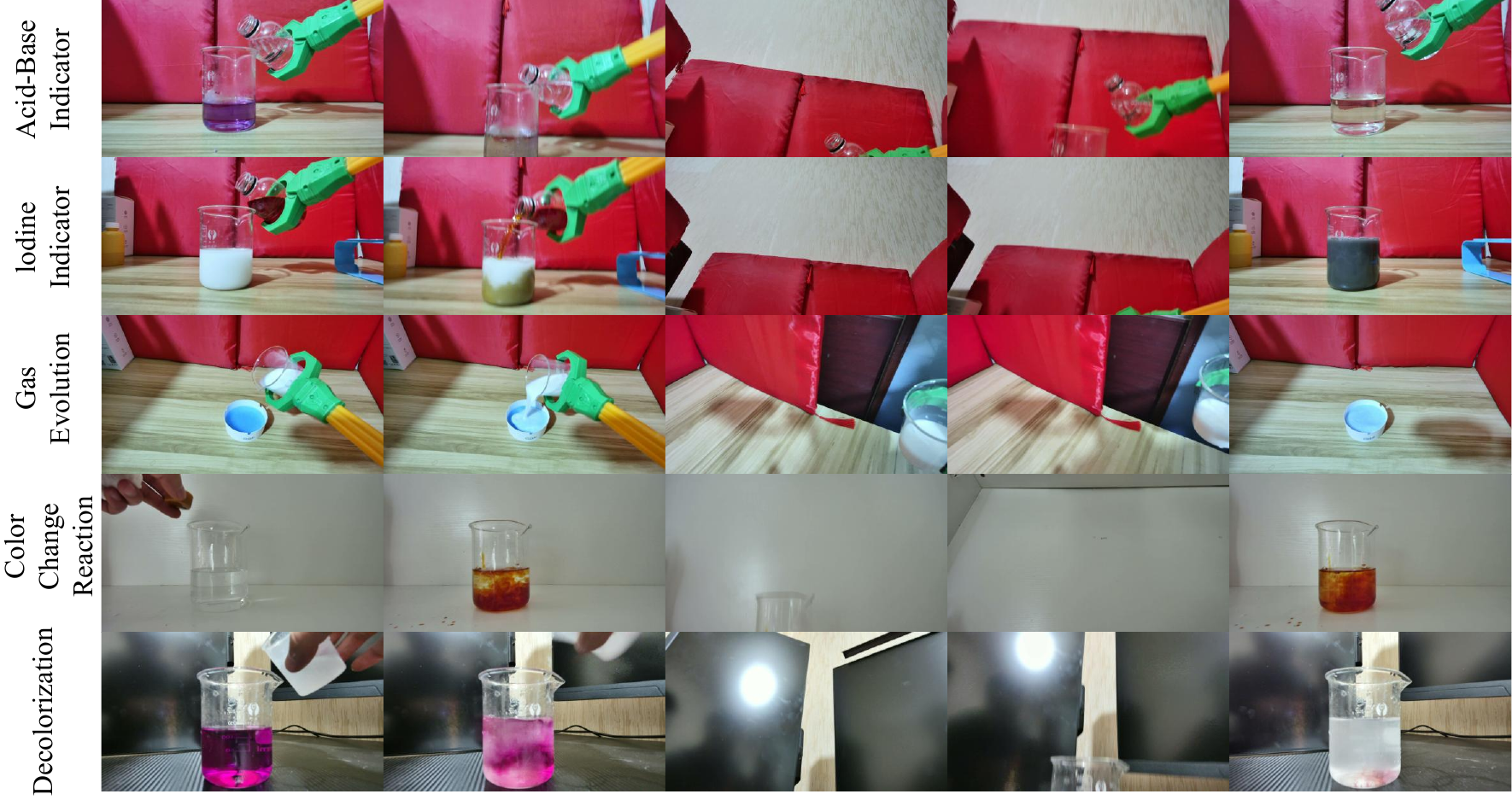}
    \caption{\textbf{Real-world examples: Chemical Reaction.} Sampled V-D-R frames showing chemical reactions such as oxidation or effervescence, where the object undergoes compositional changes during the D-phase.}
    \label{fig:sup_real4}
\end{figure*}

\begin{figure*}[t]
    \centering
    \includegraphics[width=\linewidth]{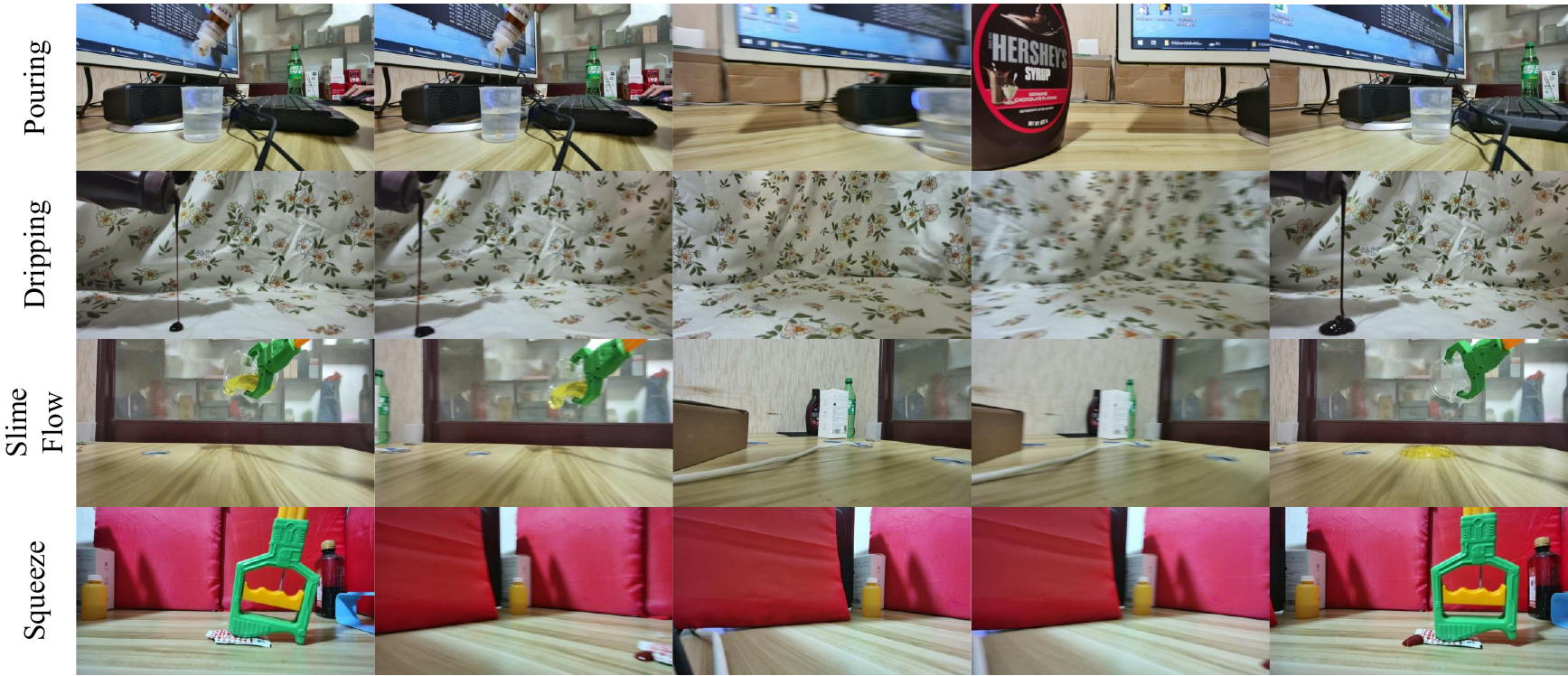}
    \caption{\textbf{Real-world examples: Viscous Flow.} Sampled V-D-R frames capturing viscous flow processes such as pouring, dripping, and slime deformation, where fluid dynamics govern the state change.}
    \label{fig:sup_real5}
\end{figure*}

\begin{figure*}[t]
    \centering
    \includegraphics[width=\linewidth]{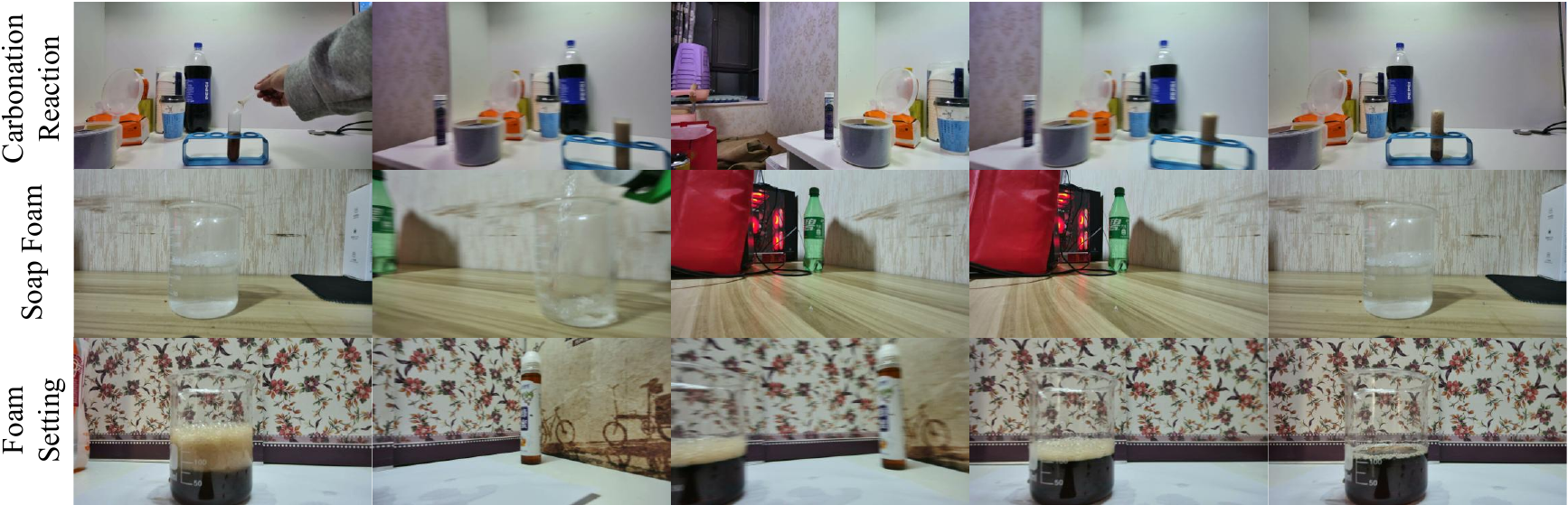}
    \caption{\textbf{Real-world examples: Bubble \& Foam.} Sampled V-D-R frames showing foam settling, soap bubble evolution, and carbonation reactions, where transient structures form and collapse over time.}
    \label{fig:sup_real6}
\end{figure*}

\begin{figure*}[t]
    \centering
    \includegraphics[width=\linewidth]{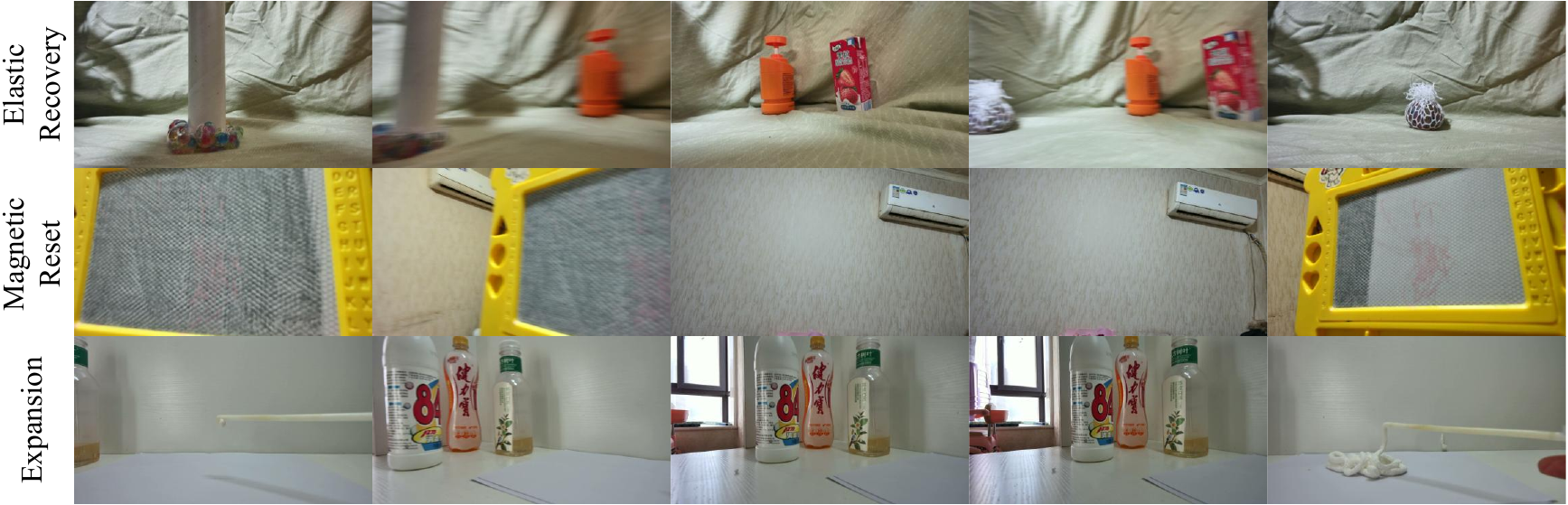}
    \caption{\textbf{Real-world examples: Physical Deformation.} Sampled V-D-R frames depicting mechanical deformations such as crushing, tearing, or bending, where the object's geometry changes through applied force.}
    \label{fig:sup_real7}
\end{figure*}

\minisection{Real-World Data.}
\cref{fig:sup_real1,fig:sup_real2,fig:sup_real3,fig:sup_real4,fig:sup_real5,fig:sup_real6,fig:sup_real7}
present examples from the real-world subset, organized by the seven state-change categories. During the D-phase, the camera pans away while the physical transformation occurs off-screen.

\subsection{More Qualitative Results}
\label{sec:more-qualitative}

\begin{figure*}[t]
    \centering
    \includegraphics[width=\linewidth]{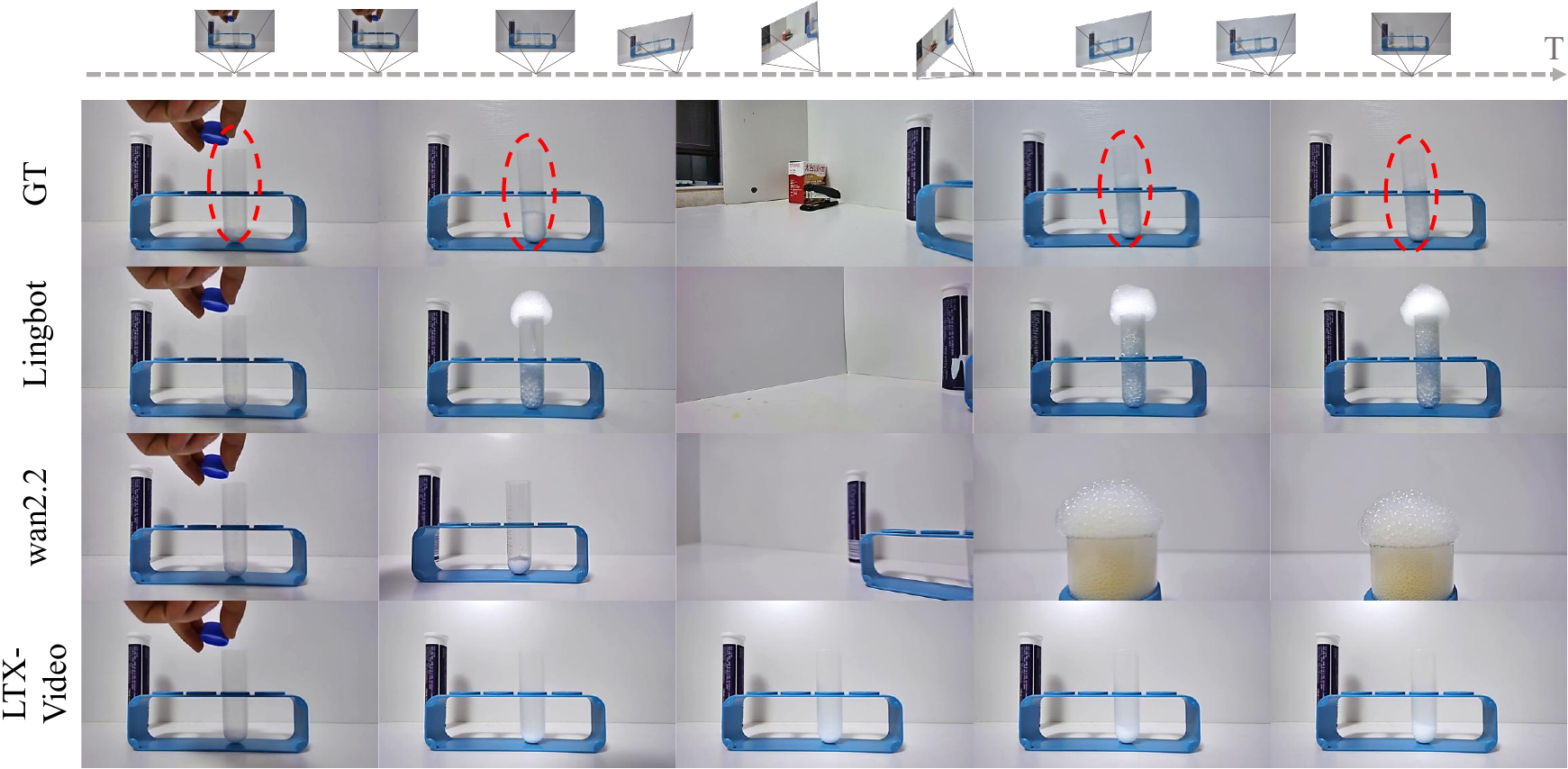}
    \caption{Qualitative comparison on a real-world clip. The camera
    trajectory is shown above, with sampled frames from LingBot,
    Wan2.2, and LTX-Video.}
    \label{fig:sup_qual1}
\end{figure*}

\begin{figure*}[t]
    \centering
    \includegraphics[width=\linewidth]{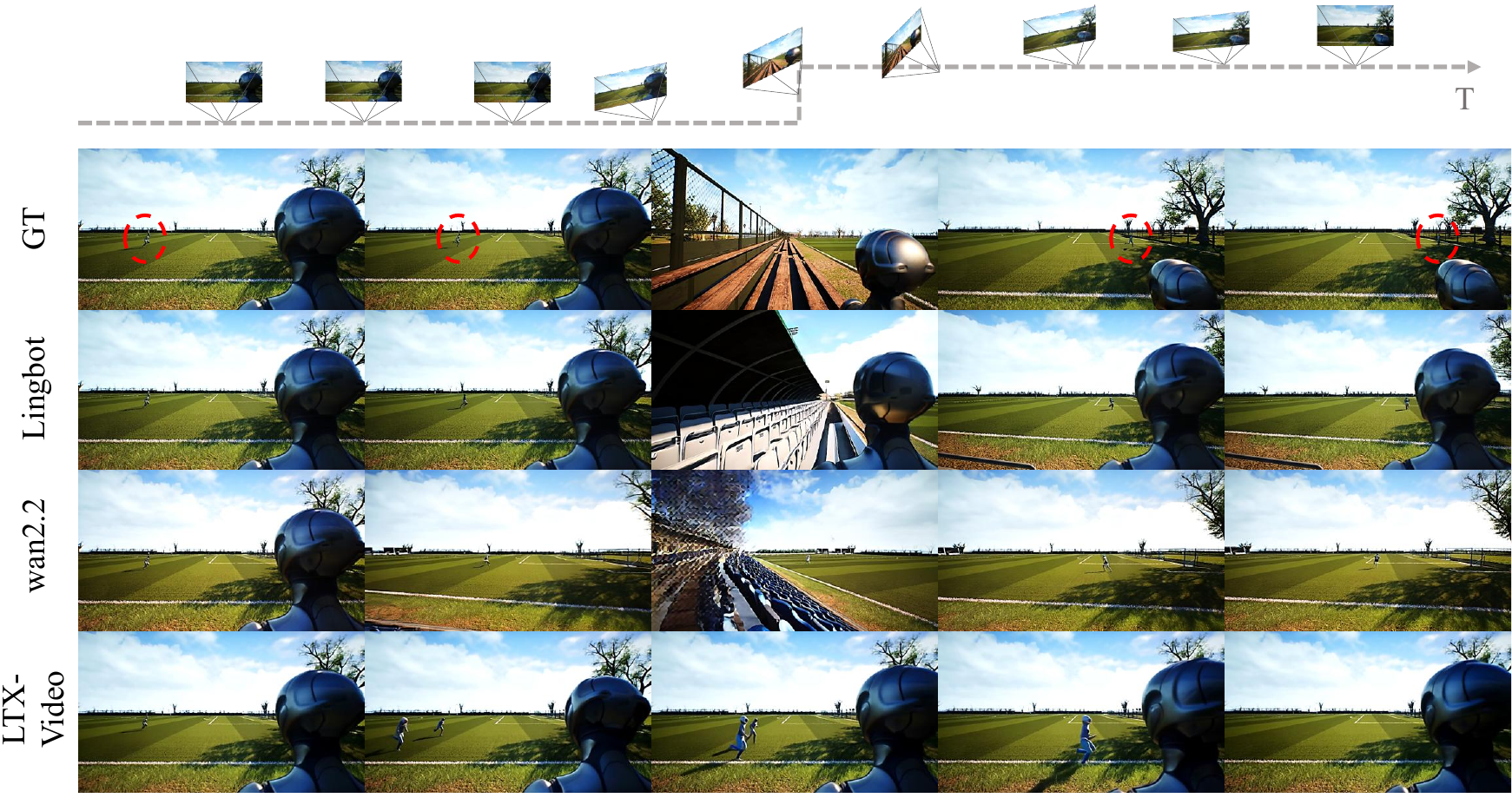}
    \caption{Qualitative comparison on a synthetic clip. The camera
    pans away from and returns to the scene, with sampled frames from
    LingBot, Wan2.2, and LTX-Video.}
    \label{fig:sup_qual2}
\end{figure*}

\begin{figure*}[t]
    \centering
    \includegraphics[width=\linewidth]{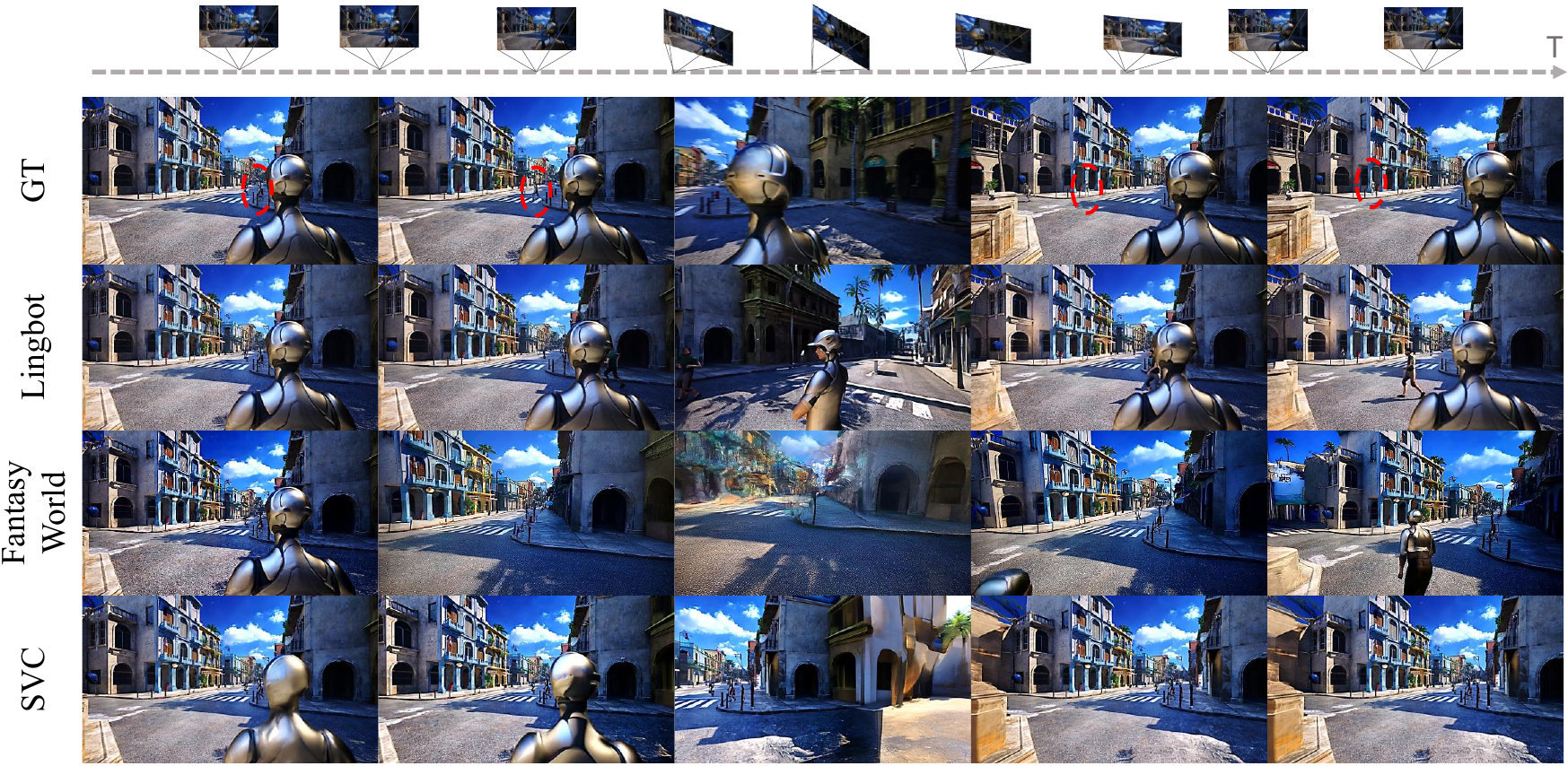}
    \caption{Qualitative comparison on a synthetic clip. The camera
    trajectory is shown above, with sampled frames from LingBot,
    FantasyWorld, and SVC.}
    \label{fig:sup_qual3}
\end{figure*}

\begin{figure*}[t]
    \centering
    \includegraphics[width=\linewidth]{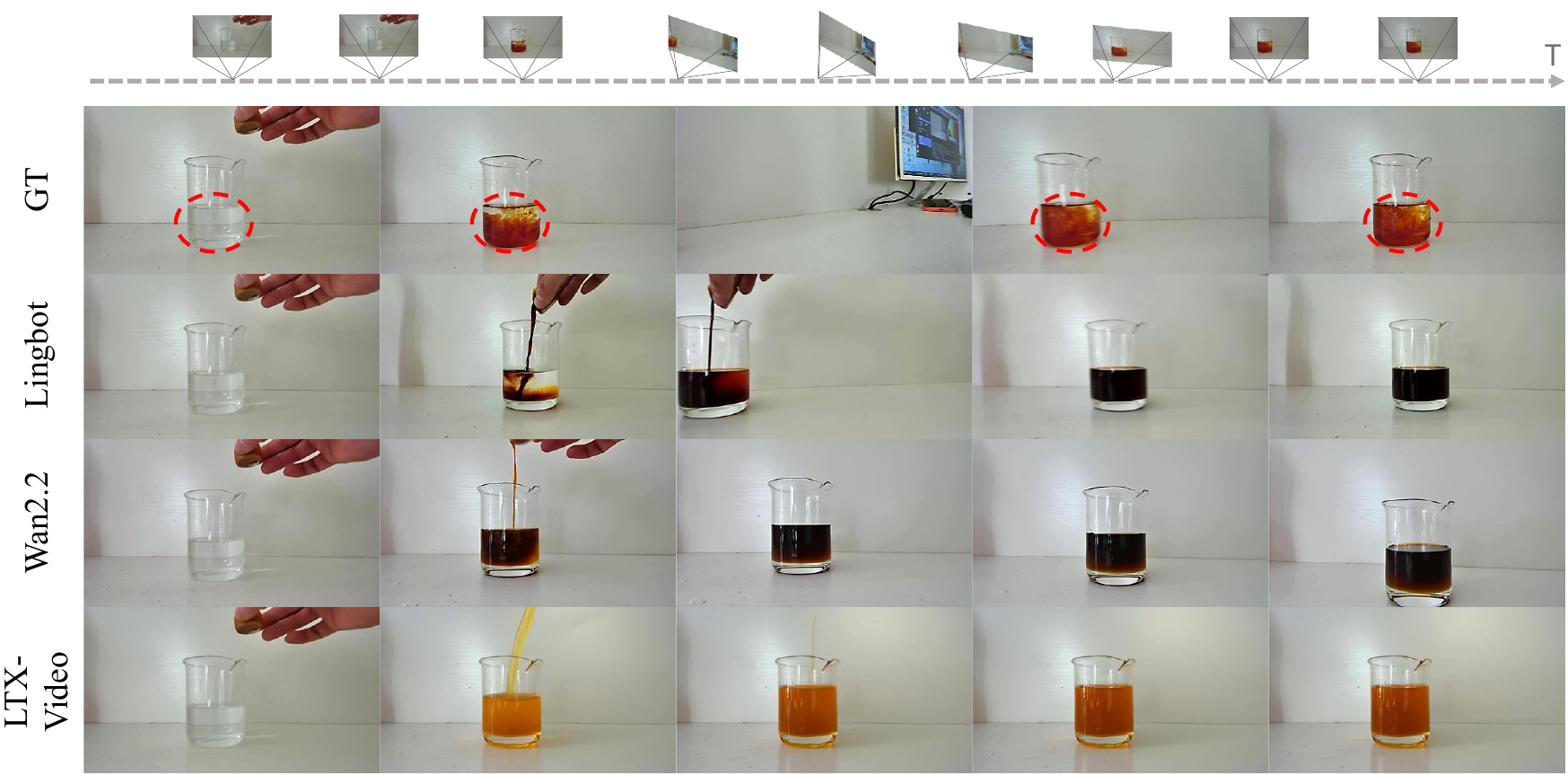}
    \caption{Qualitative comparison on a real-world clip. The camera
    departs and returns while the physical state change progresses,
    with sampled frames from LingBot, Wan2.2, and LTX-Video.}
    \label{fig:sup_qual4}
\end{figure*}

\begin{figure*}[t]
    \centering
    \includegraphics[width=\linewidth]{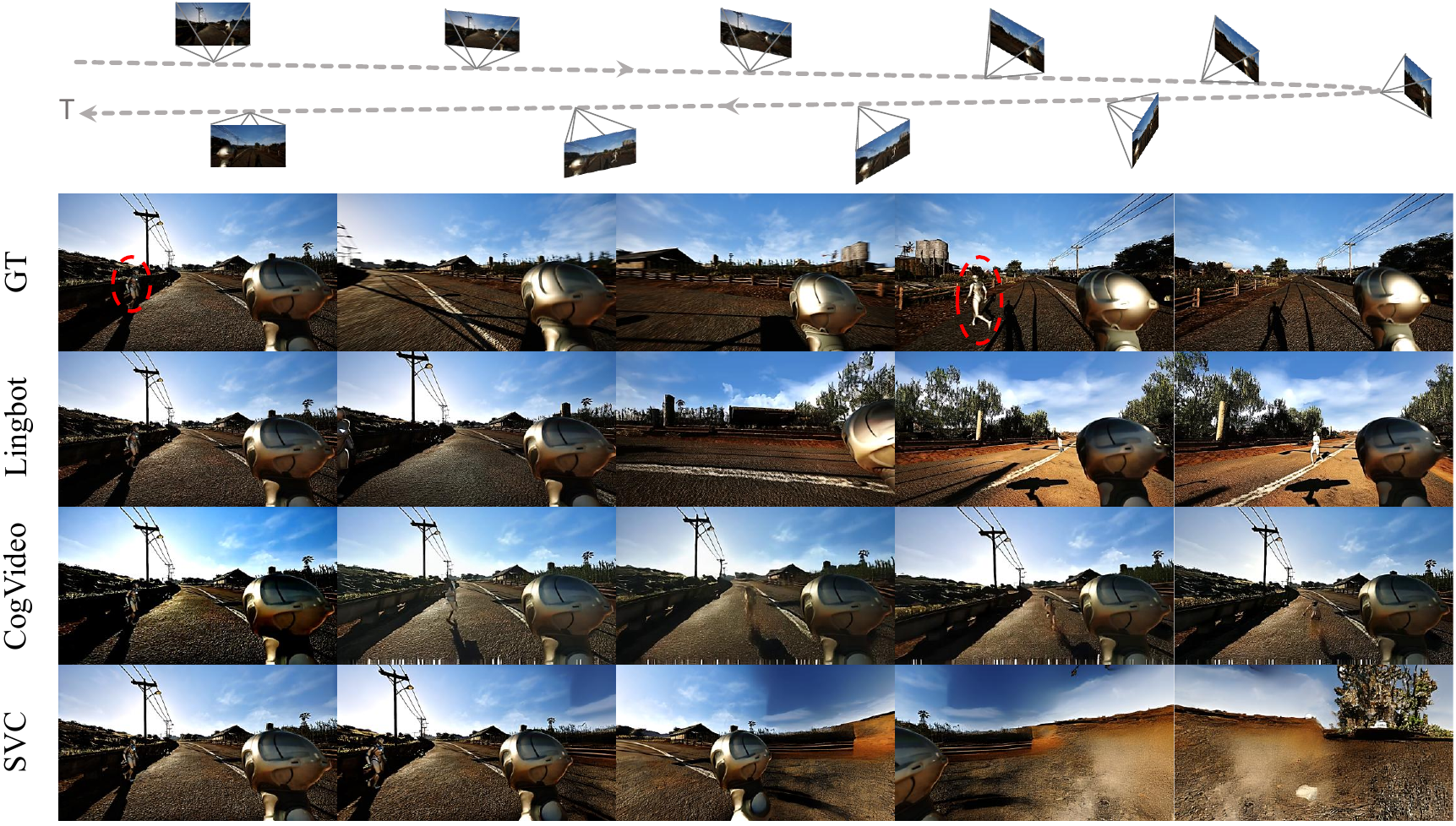}
    \caption{Qualitative comparison on a synthetic clip. The camera
    trajectory is shown above, with sampled frames from LingBot,
    CogVideoX, and SVC below.}
    \label{fig:sup_qual5}
\end{figure*}

\begin{figure*}[t]
    \centering
    \includegraphics[width=\linewidth]{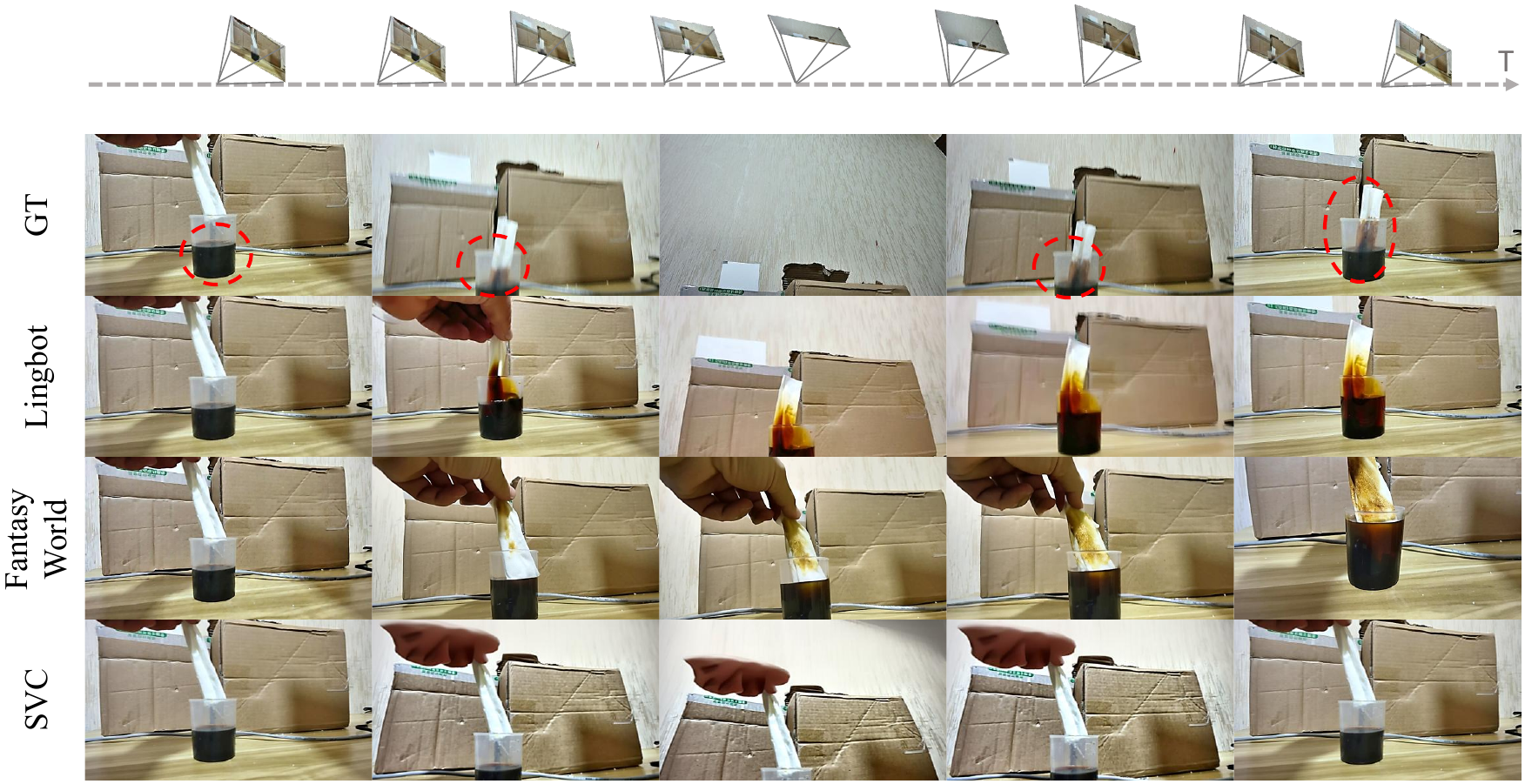}
    \caption{Qualitative comparison on a real-world clip. The camera
    departs and returns while the physical state change progresses,
    with sampled frames from LingBot, FantasyWorld, and SVC.}
    \label{fig:sup_qual6}
\end{figure*}

\cref{fig:sup_qual1,fig:sup_qual2,fig:sup_qual3,fig:sup_qual4,fig:sup_qual5,fig:sup_qual6}
show additional qualitative comparisons.
Each figure visualizes the camera trajectory alongside sampled frames from V, D, and R phases for a subset of models.

\subsection{Additional Radar Visualizations}
\label{sec:supp-radar}
\begin{figure*}[htbp]
  \centering
  \begin{subfigure}[b]{0.48\linewidth}
    \centering
    \includegraphics[width=\linewidth]{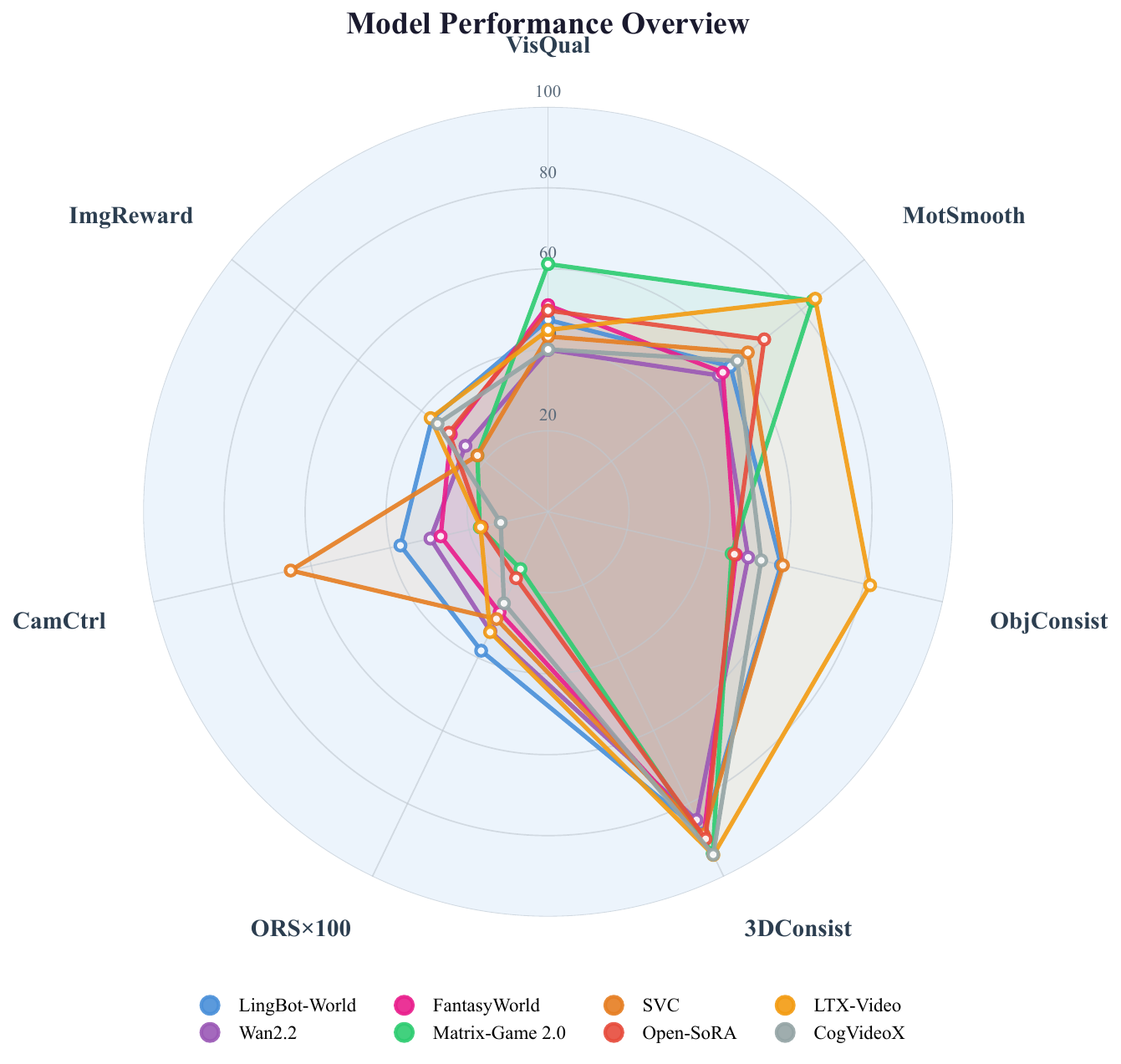}
    \caption{\textbf{Overall radar comparison.} Per-model scores across the main evaluation dimensions.}
    \label{fig:sup_radar_chart}
  \end{subfigure}
  \hfill
  \begin{subfigure}[b]{0.48\linewidth}
    \centering
    \includegraphics[width=\linewidth]{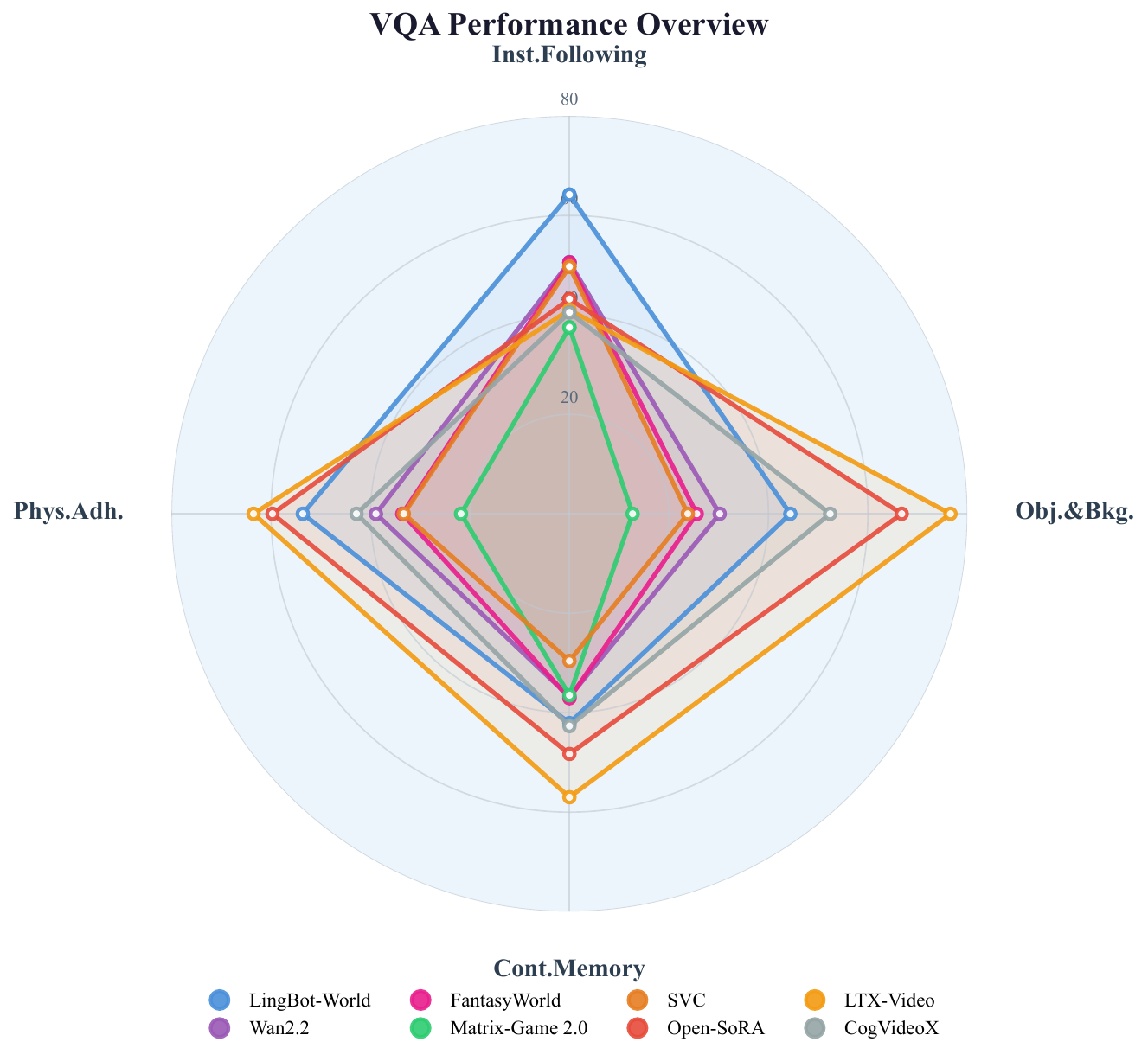}
    \caption{\textbf{Fine-grained VQA radar analysis.} Per-model breakdown across four VQA dimensions.}
    \label{fig:sup_radar_vqa}
  \end{subfigure}
  \caption{\textbf{Radar visualizations for detailed VQA evaluation.}
  The two panels provide complementary summaries: a high-level
  cross-dimension view and a fine-grained VQA-focused breakdown.}
  \label{fig:sup_radar_combined}
\end{figure*}

We provide two radar-plot visualizations to summarize VQA performance
from complementary perspectives. \cref{fig:sup_radar_chart} compares
overall performance across the key evaluation dimensions, while
\cref{fig:sup_radar_vqa} presents a fine-grained VQA-focused breakdown
of model behavior.

\FloatBarrier
\clearpage

\subsection{Ablation Studies}
\label{sec:ablation}

We conduct six ablation and diagnostic studies to validate the
robustness of our evaluation metrics and the necessity of our design choices. These studies use the models and subsets specified in each subsection and are intended as controlled diagnostics rather than a reproduction of the complete ten-model leaderboard.

\subsubsection{ORS Robustness Analysis}
\label{sec:ablation_ors}

The Object Revisit Score (ORS) relies on SAM-3 text-prompted segmentation to detect whether a target object reappears in the R-phase. We verify that ORS is stable under perturbations to (a)~the coverage-filtering thresholds used to discard spurious masks, and (b)~the text prompt formulation.

\paragraph{Coverage threshold sweep.}
ORS filters SAM-3 masks by image-area coverage $[\text{cov}_{\min}, \text{cov}_{\max}]$ to exclude noise (tiny masks) and background (overly large masks). We sweep eight threshold configurations on 30 clips for two representative models.
Table~\ref{tab:ors_threshold} reports the results: for LingBot-World the mean ORS varies by only 0.013 across all configurations (0.456--0.469), and for StableVirtualCamera by 0.020 (0.325--0.345).

\begin{table}[t]
\centering
\caption{\textbf{ORS sensitivity to coverage thresholds.} Mean ORS ($\pm$ std) across 30 clips under different mask-coverage filter ranges $[\text{cov}_{\min}, \text{cov}_{\max}]$. The default setting is highlighted.}
\label{tab:ors_threshold}
\setlength{\tabcolsep}{4pt}
\renewcommand{\arraystretch}{1}
\begin{adjustbox}{width=0.8\linewidth}
\begin{tabular}{cc cc}
\toprule
\textbf{cov\textsubscript{min} (\%)} & \textbf{cov\textsubscript{max} (\%)} & \textbf{LingBot-World} & \textbf{SVC} \\
\midrule
0.01 & 30 & 0.469 $\pm$ 0.391 & 0.345 $\pm$ 0.404 \\
0.03 & 40 & 0.462 $\pm$ 0.396 & 0.345 $\pm$ 0.404 \\
\rowcolor{violet!10} 0.05 & 50 & 0.462 $\pm$ 0.396 & 0.345 $\pm$ 0.404 \\
0.05 & 60 & 0.462 $\pm$ 0.396 & 0.345 $\pm$ 0.404 \\
0.05 & 70 & 0.462 $\pm$ 0.396 & 0.345 $\pm$ 0.404 \\
0.10 & 50 & 0.462 $\pm$ 0.396 & 0.326 $\pm$ 0.411 \\
0.10 & 70 & 0.462 $\pm$ 0.396 & 0.326 $\pm$ 0.411 \\
0.20 & 50 & 0.456 $\pm$ 0.401 & 0.325 $\pm$ 0.412 \\
\bottomrule
\end{tabular}
\end{adjustbox}
\end{table}

\paragraph{Prompt variation sweep.}
We test five prompt formulations: the original subject phrase extracted from the scene description, and four rephrasings.
Table~\ref{tab:ors_prompt} shows that semantically equivalent prompts (\emph{original} vs.\ \emph{the $\langle$subject$\rangle$}) yield nearly identical ORS, while truncated prompts (\emph{short}: first word only) degrade substantially. This gap indicates that ORS captures semantic object identity rather than low-level pattern matching.

\begin{table}[t]
\centering
\caption{\textbf{ORS sensitivity to text prompt formulation.} Mean ORS ($\pm$ std) across 30 clips under different prompt templates for SAM-3.}
\label{tab:ors_prompt}
\setlength{\tabcolsep}{4pt}
\renewcommand{\arraystretch}{1}
\begin{adjustbox}{width=0.8\linewidth}
\begin{tabular}{l cc}
\toprule
\textbf{Prompt style} & \textbf{LingBot-World} & \textbf{SVC} \\
\midrule
\rowcolor{violet!10} Original subject phrase & 0.462 $\pm$ 0.396 & 0.345 $\pm$ 0.404 \\
``the $\langle$subject$\rangle$'' & 0.459 $\pm$ 0.366 & 0.328 $\pm$ 0.406 \\
``detect $\langle$subject$\rangle$ in the scene'' & 0.284 $\pm$ 0.347 & 0.241 $\pm$ 0.313 \\
``a photo of $\langle$subject$\rangle$'' & 0.168 $\pm$ 0.304 & 0.122 $\pm$ 0.282 \\
First word only & 0.030 $\pm$ 0.029 & 0.015 $\pm$ 0.011 \\
\bottomrule
\end{tabular}
\end{adjustbox}
\end{table}

\paragraph{Stratification by object size and reappearance angle.}
We further stratify ORS by GT mask area (small $<2\%$, medium $2$--$10\%$, large $>10\%$) and by camera rotation at the reappearance frame.
Larger objects yield higher ORS (LingBot-World: large 0.67, medium 0.45, small 0.31), consistent with the intuition that larger targets are easier for the model to regenerate faithfully. Objects reappearing after extreme rotations ($>120^{\circ}$) yield near-zero ORS for StableVirtualCamera (0.005), which we attribute to generation failures at large viewpoint changes rather than detector limitations, since SAM-3 achieves a 100\% detection rate on all clips.

\subsubsection{Motion-Gated Evaluation}
\label{sec:ablation_motion}

A potential confound in camera-controllable generation benchmarks is \emph{camera inactivity}: a model that ignores the camera trajectory and produces a near-static video may receive artificially high pixel-fidelity scores.
To disentangle generation quality from camera compliance, we re-evaluate all metrics on subsets filtered by the total GT camera rotation magnitude.

Table~\ref{tab:motion_gated} reports results at the $\geq 90^{\circ}$ threshold (80 clips per model).
The most notable finding is the trade-off between camera tracking and object permanence: StableVirtualCamera reaches 92.43 Camera Controllability yet drops to 0.012 ORS, while LingBot-World balances both axes (CamCtrl 75.04, ORS 0.281). This trade-off is invisible in the aggregate evaluation and can only be surfaced by jointly reporting both metrics under controlled camera motion, underscoring the need for MemoBench's multi-dimensional protocol.
TI2V models (CogVideoX, LTX-Video) show stable scores across thresholds, as expected for models that do not condition on camera poses.

\begin{table}[t]
\centering
\caption{\textbf{Motion-gated evaluation} on clips with $\geq 90^{\circ}$ total GT camera rotation.}
\label{tab:motion_gated}
\setlength{\tabcolsep}{3pt}
\renewcommand{\arraystretch}{1.15}
\begin{adjustbox}{width=\linewidth}
\begin{tabular}{l r rrrr rr}
\toprule
\textbf{Model} & \textbf{n} & \textbf{CamCtrl} $\uparrow$ & \textbf{ORS} $\uparrow$ & \textbf{PSNR} $\uparrow$ & \textbf{SSIM} $\uparrow$ & \textbf{ObjConsist} $\uparrow$ & \textbf{3DConsist} $\uparrow$ \\
\midrule
SVC              & 80 & \textbf{92.43} & 0.012 & 12.56 & 0.34 & 38.33 & 91.42 \\
LingBot-World    & 80 & 75.04 & \textbf{0.281} & 12.12 & 0.33 & 50.49 & 88.60 \\
Wan2.2           & 80 & 65.02 & 0.057 & 11.05 & 0.27 & 22.29 & 83.78 \\
LTX-Video        & 80 & 51.08 & 0.248 & 12.13 & 0.33 & \textbf{84.50} & 91.79 \\
Open-SoRA        & 80 & 50.49 & 0.093 & 12.81 & 0.31 & 65.56 & 92.31 \\
Matrix-Game2     & 80 & 46.81 & 0.008 & 12.49 & 0.27 & 39.99 & \textbf{95.26} \\
CogVideoX        & 45 & 83.32 & 0.211 & 11.05 & 0.31 & 67.62 & 95.48 \\
FantasyWorld     & 80 & 70.81 & 0.062 & 11.43 & 0.27 & 30.92 & 85.98 \\
\bottomrule
\end{tabular}
\end{adjustbox}
\end{table}

\subsubsection{Per-Phase Fidelity Breakdown}
\label{sec:ablation_phase}

The V-D-R paradigm enables phase-aware evaluation. We separately compute pixel-fidelity metrics (PSNR, SSIM, LPIPS) for the V-phase (Visible) and R-phase (Reappear) and report the fidelity drop $\Delta$ upon object reappearance.

Table~\ref{tab:per_phase_fidelity} reveals consistent R-phase degradation across all eight models. LingBot-World suffers the largest PSNR drop ($\Delta = 5.24$\,dB) despite having the highest V-phase fidelity among the camera-conditioned C+TI2V models, suggesting that its generative prior does not maintain coherence across the occlusion gap. Matrix-Game2 exhibits the steepest perceptual degradation ($\Delta$SSIM $= 0.239$, $\Delta$LPIPS $= 0.302$). These phase-level differences would be masked in an aggregate fidelity score, motivating the per-phase breakdown in our protocol.
\begin{table}[t]
\centering
\caption{\textbf{Per-phase pixel fidelity breakdown.} V = Visible phase, R = Reappear phase. $\Delta$ denotes the fidelity drop upon reappearance (positive = degradation).}
\label{tab:per_phase_fidelity}
\setlength{\tabcolsep}{3.5pt}
\renewcommand{\arraystretch}{1.15}
\begin{adjustbox}{width=\linewidth}
\begin{tabular}{l rrr rrr rrr}
\toprule
 & \multicolumn{3}{c}{\textbf{PSNR} $\uparrow$} & \multicolumn{3}{c}{\textbf{SSIM} $\uparrow$} & \multicolumn{3}{c}{\textbf{LPIPS} $\downarrow$} \\
\cmidrule(lr){2-4} \cmidrule(lr){5-7} \cmidrule(lr){8-10}
\textbf{Model} & V & R & $\Delta$ & V & R & $\Delta$ & V & R & $\Delta$ \\
\midrule
SVC              & 18.68 & 14.56 & +4.12 & 0.632 & 0.483 & +0.148 & 0.295 & 0.473 & +0.178 \\
LingBot-World    & 17.96 & 12.72 & +5.24 & 0.611 & 0.420 & +0.191 & 0.309 & 0.545 & +0.236 \\
Matrix-Game2     & 17.02 & 11.90 & +5.12 & 0.527 & 0.289 & +0.239 & 0.335 & 0.637 & +0.302 \\
Wan2.2           & 16.20 & 12.81 & +3.38 & 0.552 & 0.421 & +0.131 & 0.374 & 0.581 & +0.207 \\
FantasyWorld     & 16.09 & 11.93 & +4.16 & 0.525 & 0.376 & +0.149 & 0.406 & 0.628 & +0.222 \\
LTX-Video        & 15.91 & 12.74 & +3.16 & 0.543 & 0.420 & +0.123 & 0.363 & 0.543 & +0.180 \\
Open-SoRA        & 15.23 & 11.00 & +4.23 & 0.467 & 0.339 & +0.128 & 0.401 & 0.658 & +0.257 \\
CogVideoX        & 14.18 & 11.17 & +3.01 & 0.555 & 0.441 & +0.114 & 0.408 & 0.672 & +0.264 \\
\bottomrule
\end{tabular}
\end{adjustbox}
\end{table}

\subsubsection{Metric Sensitivity Analysis}
\label{sec:ablation_metric}

Each metric in our pipeline depends on hyperparameters (\eg, the fraction of DINOv2 patch tokens retained, the optical-flow outlier threshold, or the depth-sampling density). We sweep these parameters and measure Kendall's $\tau$ rank correlation against the default configuration over 200 clip--model pairs to verify that model rankings are not artifacts of a particular parameter choice.

Table~\ref{tab:metric_sensitivity} reports the results.
RAFT Motion Smoothness is perfectly rank-preserving ($\tau = 1.000$) across the entire threshold range $[0.05, 0.30]$, indicating that the outlier threshold affects absolute scores but not relative ordering.
DINOv2 Object Identity Consistency maintains $\tau \geq 0.910$ even when the top-$k$ fraction varies from 0.2 to 0.8.
Depth Anything V2 Geo3D Consistency is the most sensitive of the three, yet still achieves $\tau \geq 0.860$ across all sampling densities. 
All correlations are statistically significant ($p < 10^{-4}$).
\begin{table}[t]
\centering
\caption{\textbf{Metric sensitivity analysis.} Kendall's $\tau$ rank correlation between default and variant hyperparameter settings.}
\label{tab:metric_sensitivity}
\setlength{\tabcolsep}{2pt}
\renewcommand{\arraystretch}{1}
\begin{adjustbox}{width=0.8\linewidth}
\begin{tabular}{llc}
\toprule
\textbf{Metric} & \textbf{Parameter variant} & \textbf{Kendall's $\tau$} \\
\midrule
\multirow{5}{*}{DINOv2 ObjConsist} & top\_k = 0.2 & 0.947 \\
 & top\_k = 0.3 & 0.976 \\
 & \cellcolor{violet!10} top\_k = 0.4 (default) & \cellcolor{violet!10} 1.000 \\
 & top\_k = 0.5 & 0.978 \\
 & top\_k = 0.6 & 0.958 \\
 & top\_k = 0.8 & 0.910 \\
\midrule
\multirow{4}{*}{RAFT MotSmooth} & $\tau$ = 0.05 & 1.000 \\
 & $\tau$ = 0.10 & 1.000 \\
 & \cellcolor{violet!10} $\tau$ = 0.15 (default) & \cellcolor{violet!10} 1.000 \\
 & $\tau$ = 0.20 & 1.000 \\
 & $\tau$ = 0.30 & 1.000 \\
\midrule
\multirow{3}{*}{DepthV2 Geo3D} & n\_sample = 2 & 0.873 \\
 & \cellcolor{violet!10} n\_sample = 3 (default) & \cellcolor{violet!10} 1.000 \\
 & n\_sample = 5 & 0.900 \\
 & n\_sample = 7 & 0.860 \\
\bottomrule
\end{tabular}
\end{adjustbox}
\end{table}

\subsubsection{Camera Pose Estimation Validation}
\label{sec:ablation_pose}

Camera Controllability is derived from the Absolute Trajectory Error (ATE) between MapAnything-estimated and GT camera poses. We perform two sanity checks: (1)~verifying GT pose integrity, and (2)~examining whether the ATE distribution across models is consistent with their architectural priors.
For (1), we recompute the total rotation from raw GT \texttt{poses.npy} files and compare against the values stored in the evaluation CSVs. Across all 159 synthetic clips the Pearson correlation is $r = 1.0000$ with a maximum absolute difference below $0.01^{\circ}$.
For (2), Table~\ref{tab:pose_validation} reports the per-model ATE distribution on synthetic data. StableVirtualCamera, which directly conditions on target camera extrinsics, achieves the lowest ATE ($8.3^{\circ} \pm 6.6^{\circ}$). LingBot-World ranks second ($20.7^{\circ}$). TI2V models that receive no camera input (CogVideoX, LTX-Video, Open-SoRA) cluster near $63$--$65^{\circ}$, consistent with near-random camera trajectories.
The Spearman correlation between ATE and Camera Controllability is moderate ($\rho = -0.355$); the non-linear mapping from ATE to the coverage-based CamCtrl score accounts for this gap, since CamCtrl saturates once the ATE falls below the coverage radius.
\begin{table}[htbp]
\centering
\caption{\textbf{Camera pose estimation validation.} Per-model ATE rotation RMSE (degrees) on synthetic data, sorted by ATE. CamCtrl is reported on a 0--100 scale.}
\label{tab:pose_validation}
\setlength{\tabcolsep}{5pt}
\renewcommand{\arraystretch}{1}
\begin{adjustbox}{width=0.8\linewidth}
\begin{tabular}{lrrrr}
\toprule
\textbf{Model} & \textbf{n} & \textbf{ATE ($^{\circ}$)} $\downarrow$ & \textbf{CamCtrl} $\uparrow$ & \textbf{GT Rot ($^{\circ}$)} \\
\midrule
SVC              & 159 & 8.3 $\pm$ 6.6  & 86.10 & 93.6 \\
LingBot-World    & 159 & 20.7 $\pm$ 24.1 & 93.06 & 93.6 \\
FantasyWorld     & 159 & 34.4 $\pm$ 24.5 & 52.76 & 93.6 \\
Wan2.2           & 159 & 44.9 $\pm$ 36.0 & 66.15 & 93.6 \\
Matrix-Game2     & 159 & 58.1 $\pm$ 30.6 & 32.52 & 93.6 \\
Open-SoRA        & 159 & 63.0 $\pm$ 38.9 & 32.77 & 93.6 \\
LTX-Video        & 159 & 63.6 $\pm$ 39.1 & 32.55 & 93.6 \\
CogVideoX        &  86 & 64.5 $\pm$ 38.3 & 31.15 & 95.7 \\
\bottomrule
\end{tabular}
\end{adjustbox}
\end{table}
\subsubsection{Initial-State Conditioning vs.\ Backbone Capacity}
\label{sec:ablation_capacity}

To disentangle the effect of initial-state conditioning from backbone capacity, we conduct a controlled Wan2.2 ablation on 50 clips. The \emph{with V-frame} setting provides the first frame of the V phase as an image condition together with the text prompt and camera trajectory.
The \emph{without V-frame} setting removes this image condition and uses only text and camera conditioning.

As shown in Table~\ref{tab:wan_capacity}, providing the V-phase frame improves GT-aligned fidelity more substantially than scaling the backbone from 5B to 14B. Adding the V-frame improves PSNR by 4.2\,dB for the 5B model and 4.7\,dB for the 14B model, while also reducing LPIPS by 0.20 and 0.16, respectively. In comparison, scaling the backbone from 5B to 14B produces substantially smaller improvements under matched conditioning.

Interestingly, the 14B model without the V-frame achieves the highest ORS, Object Consistency, Motion Smoothness, and Geo3D Consistency, despite its substantially lower GT-aligned fidelity. This result shows that internally self-consistent generation does not necessarily recover the correct post-occlusion state, motivating the joint use of GT-aligned fidelity and self-consistency metrics.

\begin{table}[t]
\centering
\caption{
\textbf{Wan2.2 initial-state-conditioning and backbone-capacity
ablation} on 50 clips. ``V-frame'' denotes the first frame of the visible phase provided as an image condition.
}
\label{tab:wan_capacity}
\scriptsize
\setlength{\tabcolsep}{3pt}
\renewcommand{\arraystretch}{1.1}
\begin{adjustbox}{width=\linewidth}
\begin{tabular}{lcccccc}
\toprule
\textbf{Variant}
& \textbf{PSNR}$\uparrow$
& \textbf{LPIPS}$\downarrow$
& \textbf{ORS}$\uparrow$
& \textbf{ObjConsist}$\uparrow$
& \textbf{MotSmooth}$\uparrow$
& \textbf{3DConsist}$\uparrow$ \\
\midrule
5B w/o V-frame
& 8.70 & 0.85 & 0.27 & 47.1 & 61.7 & 90.7 \\
5B w/ V-frame
& 12.9 & 0.65 & 0.36 & 49.2 & 65.5 & 88.7 \\
14B w/o V-frame
& 8.90 & 0.74 & \textbf{0.53} & \textbf{80.7}
& \textbf{79.7} & \textbf{97.0} \\
14B w/ V-frame
& \textbf{13.6} & \textbf{0.58} & 0.21 & 45.6
& 57.3 & 83.7 \\
\bottomrule
\end{tabular}
\end{adjustbox}
\end{table}

\FloatBarrier
\clearpage

\subsection{Detailed VQA Pipeline}
\label{sec:complete-vqa}

We present the three-stage VQA pipeline used in our evaluation.
Each stage is illustrated with its prompt template, followed by
concrete filtering examples.

\newcommand{\promptcard}[2]{%
\par\smallskip
\noindent\makebox[\linewidth][c]{%
\setlength{\fboxsep}{8pt}%
\fcolorbox{black!60}{white}{%
\begin{minipage}{0.97\linewidth}
\colorbox{black!65}{\parbox{\dimexpr\linewidth-2\fboxsep\relax}{\textcolor{white}{\ttfamily\small\strut #1}}}\par\vspace{14pt}\noindent{\small #2}
\end{minipage}}}
\par\smallskip
}
\vspace{10pt}
\noindent\textbf{Stage 1: LLM Question Generation.} Given the start frame and generation prompt, we ask an LLM to generate candidate Yes/No questions (six per dimension).
\promptcard{Question Generation Prompt}{%
\textbf{System Role:}\\
You are an expert LLM judger, specializing in ``World Model'' evaluation. Your task is to generate questions used to evaluate AI-generated videos against text instructions.

\textbf{Input Data:}\\
Start Frame (Ground Truth) as attached image\\
Generation Prompt: \colorbox{yellow!30}{\texttt{\{generation\_prompt\}}}

\textbf{Task:}\\
Generate 24 Yes/No questions.

\textbf{Evaluation Dimensions \& Constraints:}\\
Generate six questions for each of the four dimensions below. We use \textbf{mixed polarity}, meaning no dimension has a fixed yes/no preference.
\begin{itemize}[nosep,leftmargin=*]
    \item \textbf{Instruction Following}: Did the video follow the requested movements and events?
    \item \textbf{Object and Background}: Is there inconsistency in subject identity or background details?
    \item \textbf{Continuity of Memory}: Does the model preserve object location/trajectory while out of frame?
    \item \textbf{Physics Adherence}: Are lighting, shadows, and motion physically plausible?
\end{itemize}

\textbf{Output Format:}\\
Output JSON with columns: [ID, Dimension, Question].
}
\clearpage
\noindent\textbf{Stage 2: Question Filtering.} Candidate questions are filtered using both ground-truth and failure-case references.

\promptcard{GT \& Failure Filtering Prompt}{%
\textbf{System Role:}\\
You are an expert LLM judger, specializing in ``World Model'' evaluation. Your task is to audit AI-generated videos against specific text instructions.

\textbf{Input Data:}\\
Test Video as attached video\\
Questions: \colorbox{yellow!30}{\texttt{\{questions\}}}

\textbf{Task:}\\
Watch the test video and answer each Yes/No question.

\textbf{Output Format:}\\
Output JSON with columns: [ID, Dimension, Question, Answer (Yes/No), Verdict (Pass/Fail), Reasoning].
}\promptcard{Revised Failure Filtering Prompt (with Hint)}{%
\textbf{System Role:}\\
You are an expert LLM judger, specializing in ``World Model'' evaluation. Your task is to audit AI-generated videos against specific text instructions and known failure hints.

\textbf{Input Data:}\\
Test Video as attached video\\
Questions: \colorbox{yellow!30}{\texttt{\{questions\}}}\\
Hint: \colorbox{yellow!30}{\texttt{\{hint\}}}

\textbf{Tasks:}
\begin{enumerate}[nosep,leftmargin=*]
    \item Answer the Yes/No questions for the test video.
    \item Audit these answers against known failures and remove unstable questions.
\end{enumerate}

\textbf{Output Format:}\\
Output JSON with columns: [ID, Dimension, Question, Answer (Yes/No), Verdict (Pass/Fail), Reasoning].
}

\noindent\textbf{Example Legend} In the examples below, \textcolor{green!60!black}{\ding{51}} denotes a question retained after filtering; \textcolor{red}{\ding{55}\,\scriptsize Failure} denotes a question removed due to instability under failure-case checking; and \textcolor{red}{\ding{55}\,\scriptsize GT} denotes a question removed because it is inconsistent with the GT reference.

\noindent\textbf{Example: Nordic \#001} (5/8 questions remain after filtering).
\begin{itemize}[nosep,leftmargin=*]
    \item \textit{Instruction Following} --- (1)~Does the observer re-encounter the subject after completing the U-turn? \textcolor{green!60!black}{\ding{51}} ~(2)~Does the observer execute a U-turn after the subject has exited the field of view? \textcolor{red}{\ding{55}\,\scriptsize Failure}
    \item \textit{Object \& Background} --- (1)~Does the subject maintain its silver robotic appearance throughout? \textcolor{green!60!black}{\ding{51}} ~(2)~Does the Nordic architecture maintain its structural details during camera rotation? \textcolor{green!60!black}{\ding{51}}
    \item \textit{Continuity of Memory} --- (1)~Does the street layout change its configuration after the observer turns around? \textcolor{green!60!black}{\ding{51}} ~(2)~Is the subject in a logically consistent position when the observer turns back? \textcolor{red}{\ding{55}\,\scriptsize Failure}
    \item \textit{Physics Adherence} --- (1)~Do shadows move realistically as the observer changes perspective? \textcolor{green!60!black}{\ding{51}} ~(2)~Does the subject's walking speed remain consistent and natural? \textcolor{red}{\ding{55}\,\scriptsize Failure}
\end{itemize}

\noindent\textbf{Example: Zen Garden \#005} (4/8 questions remain after filtering).
\begin{itemize}[nosep,leftmargin=*]
    \item \textit{Instruction Following} --- (1)~Does the observer successfully perform a U-turn and see the person again? \textcolor{green!60!black}{\ding{51}} ~(2)~Does the person continue moving consistently after the fox completes the turn? \textcolor{red}{\ding{55}\,\scriptsize Failure}
    \item \textit{Object \& Background} --- (1)~Is the Zen Garden aesthetic and landscape preserved throughout? \textcolor{green!60!black}{\ding{51}} ~(2)~Does the ground texture flicker or disappear as the fox runs? \textcolor{red}{\ding{55}\,\scriptsize GT}
    \item \textit{Continuity of Memory} --- (1)~Is the character's walking animation continuous even off-focus? \textcolor{green!60!black}{\ding{51}} ~(2)~Does the person reappear at a location consistent with their original trajectory? \textcolor{red}{\ding{55}\,\scriptsize Failure}
    \item \textit{Physics Adherence} --- (1)~Do shadows cast by trees remain in a fixed orientation relative to the sun? \textcolor{green!60!black}{\ding{51}} ~(2)~Does the foliage jitter unnaturally as the camera moves past it? \textcolor{red}{\ding{55}\,\scriptsize GT}
\end{itemize}

\vspace{10pt}
\noindent\textbf{Stage 3: Final Evaluation.} The filtered question bank is applied to each test video. A VLM answers the remaining questions, and per-dimension scores are aggregated from binary verdicts.

\promptcard{Judge Prompt}{%
\textbf{System Role:}\\
You are an expert judge specialized in evaluating answer correctness for world-model outputs.

\textbf{Input Data:}\\
Test Video as attached video (generated output)\\
Evaluation Questions: \colorbox{yellow!30}{\texttt{\{filtered\_questions\}}}

\textbf{Task:}\\
Watch the test video and answer each question with a final Yes/No decision and reasoning.

\textbf{Output Format:}\\
Output JSON with columns: [ID, Dimension, Question, Answer (Yes/No), Verdict (Pass/Fail), Reasoning].
}

\FloatBarrier

\subsection{Failure Analysis}
\label{sec:failure-analysis}

We further analyze the failure modes of LingBot-World on the synthetic
and real-world subsets of \NAME. We categorize the observed failures
into six types: object disappearance, identity drift, state reset,
teleportation, background hallucination, and camera drift.
The categories are non-exclusive, and a single generated sequence may
exhibit multiple failure types. This analysis is intended as a case
study of recurring model failures rather than a comparison across
different models.

\begin{table}[t]
\centering
\caption{
\textbf{Failure taxonomy for LingBot-World.}
The reported counts are non-exclusive, since one generated sequence
may exhibit multiple failure types.
}
\label{tab:failure_taxonomy}
\scriptsize
\setlength{\tabcolsep}{2.5pt}
\renewcommand{\arraystretch}{1.1}
\begin{adjustbox}{width=\linewidth}
\begin{tabular}{lcccccc}
\toprule
\textbf{Split}
& \textbf{Obj. vanish}
& \textbf{Id. drift}
& \textbf{State reset}
& \textbf{Teleport}
& \textbf{Bkg. halluc.}
& \textbf{Cam. drift} \\
\midrule
Synthetic & 85 & 49 & 5 & 54 & 131 & 72 \\
Real      & 15 & 83 & 2 & 13 & 57  & 35 \\
\bottomrule
\end{tabular}
\end{adjustbox}
\end{table}

The two subsets exhibit different dominant failure patterns.
Background hallucination is the most frequent failure on the synthetic
subset, followed by object disappearance and camera drift. In contrast,
identity drift is the most frequent failure on the real-world subset,
where the target object may undergo a physical-state change while
outside the field of view. These results indicate that memory failures
extend beyond object disappearance and also affect object identity,
scene layout, physical state, and camera execution.

The failure categories are reflected by complementary components of
our evaluation protocol. Object disappearance is primarily captured by
ORS; identity drift by Object Consistency and Object \& Background VQA;
state reset by R-phase GT-aligned fidelity and clip-specific VQA;
teleportation by Motion Smoothness and VQA; background hallucination
by whole-frame fidelity, 3D Consistency, and Object \& Background VQA;
and camera drift by Camera Controllability.

\end{document}